\def\year{2019}\relax
\documentclass[letterpaper]{article} 
\usepackage{aaai19}  
\usepackage{times}  
\usepackage{helvet}  
\usepackage{courier}  
\usepackage{graphicx}  
\usepackage{algorithm}
\usepackage{algpseudocode}
\usepackage{amssymb}
\usepackage{booktabs}
\usepackage{cite}
\usepackage{multirow}
\usepackage{placeins}
\usepackage{subcaption}
\usepackage{stfloats}
\usepackage[dvipsnames]{xcolor}
\PassOptionsToPackage{hyphens}{url}\usepackage[colorlinks, citecolor=PineGreen, urlcolor=NavyBlue, linkcolor=Brown]{hyperref}

\algnewcommand{\LComment}[1]{\Statex\(\triangleright\) #1}
\usepackage{xspace}
\makeatletter
\DeclareRobustCommand\onedot{\futurelet\@let@token\@onedot}
\def\@onedot{\ifx\@let@token.\else.\null\fi\xspace}
\def\eg{\emph{e.g}\onedot}

\def\ie{\emph{i.e}\onedot}

\def\etc{\emph{etc}\onedot}

\makeatother

\makeatletter
\def\blindfootnote{\xdef\@thefnmark{}\@footnotetext}
\makeatother

\begin{document}
%
\title{Regularized Evolution for Image Classifier Architecture Search}
\author{Esteban Real\textsuperscript{\textasteriskcentered}\textsuperscript{\textdagger} \and Alok Aggarwal\textsuperscript{\textdagger} \and Yanping Huang\textsuperscript{\textdagger} \and Quoc V. Le\\
Google Brain, Mountain View, California, USA\\
\textsuperscript{\textdagger}Equal contribution. \textsuperscript{\textasteriskcentered}Correspondence: ereal@google.com
}
\maketitle
\begin{abstract}
The effort devoted to hand-crafting neural network image classifiers has motivated the use of architecture search to discover them automatically. Although evolutionary algorithms have been repeatedly applied to neural network topologies, the image classifiers thus discovered have remained inferior to human-crafted ones. Here, we evolve an image classifier---\mbox{\textit{AmoebaNet-A}}---that surpasses hand-designs for the first time. To do this, we modify the tournament selection evolutionary algorithm by introducing an age property to favor the younger genotypes. Matching size, \mbox{AmoebaNet-A} has comparable accuracy to current state-of-the-art ImageNet models discovered with more complex architecture-search methods. Scaled to larger size, \mbox{AmoebaNet-A} sets a new state-of-the-art 83.9\% \mbox{top-1} / 96.6\% \mbox{top-5} ImageNet accuracy. In a controlled comparison against a well known reinforcement learning algorithm, we give evidence that evolution can obtain results faster with the same hardware, especially at  the earlier stages of the search. This is relevant when fewer compute resources are available. Evolution is, thus, a simple method to effectively discover high-quality architectures.
\end{abstract}

\section{Introduction}

\blindfootnote{Accepted for publication at AAAI 2019, the Thirty-Third AAAI Conference on Artificial Intelligence.}
\blindfootnote{A brief talk from Nov 2018 summarizes this paper at \url{https://www.youtube.com/watch?v=MqYHo7BVzoE}}

Until recently, most state-of-the-art image classifier architectures have been manually designed by human experts \cite{krizhevsky2012imagenet,szegedy2015going,he2016deep,huang2016densely,hu2017squeeze}. To speed up the process, researchers have looked into automated methods \cite{baker2016designing,zoph2016neural,miikkulainen2017evolving,real2017large,xie2017genetic,suganuma2017genetic,liu2017progressive,pham2018faster}. These methods are now collectively known as \textit{architecture-search algorithms}. A traditional approach is \textit{neuro-evolution of topologies} \cite{miller1989designing,angeline1994evolutionary,stanley2002evolving}. Improved hardware now allows scaling up evolution to produce high-quality image classifiers \cite{real2017large,xie2017genetic,liu2017hierarchical}. Yet, the architectures produced by evolutionary algorithms / genetic programming have not reached the accuracy of those directly designed by human experts. Here we evolve image classifiers that surpass hand-designs.

To do this, we make two additions to the standard evolutionary process. First, we propose a change to the well-established \textit{tournament selection} evolutionary algorithm \cite{goldberg1991comparative} that we refer to as \textit{aging} evolution or \textit{regularized} evolution. Whereas in tournament selection, the best genotypes (architectures) are kept, we propose to associate each genotype with an age, and bias the tournament selection to choose the younger genotypes. We will show that this change turns out to make a difference. The connection to regularization will be clarified in the Discussion section. Second, we implement the simplest set of mutations that would allow evolving in the NASNet search space~\cite{zoph2017learning}. This search space associates convolutional neural network architectures with small directed graphs in which vertices represent hidden states and labeled edges represent common network operations (such as convolutions or pooling layers). Our mutation rules only alter architectures by randomly reconnecting the origin of edges to different vertices and by randomly relabeling the edges, covering the full search space.

Searching in the NASNet space allows a controlled comparison between evolution and the original method for which it was designed, reinforcement learning (RL). Thus, this paper presents the first comparative case study of architecture-search algorithms for the image classification task. Within this case study, we will demonstrate that evolution can attain similar results with a simpler method, as will be shown in the Discussion section. In particular, we will highlight that in all our experiments evolution searched faster than RL and random search, especially at the earlier stages, which is important when experiments cannot be run for long times due to compute resource limitations.

Despite its simplicity, our approach works well in our benchmark against RL. It also evolved a high-quality model, which we name \mbox{\textit{AmoebaNet-A}}. This model is competitive with the best image classifiers obtained by any other algorithm today at similar sizes (82.8\% top-1 / 96.1\% top-5 ImageNet accuracy). When scaled up, it sets a new state-of-the-art accuracy (83.9\% \mbox{top-1} / 96.6\% \mbox{top-5} ImageNet accuracy)\footnote{After our submission, a recent preprint has further scaled up and retrained \mbox{AmoebaNet-A} to reach 84.3\% \mbox{top-1} / 97.0\% \mbox{top-5} ImageNet accuracy \cite{huang2018gpipe}.}.

\section{Related Work}

Review papers provide informative surveys of earlier \cite{yao1999evolving,floreano2008neuroevolution} and more recent \cite{elsken2018neural} literature on image classifier architecture search, including successful RL studies \cite{zoph2016neural,baker2016designing,zoph2017learning,liu2017progressive,zhong2017practical,cai2017reinforcement} and evolutionary studies like those mentioned in the Introduction. Other methods have also been applied: cascade-correlation \cite{fahlman1990cascade}, boosting \cite{cortes2016adanet}, hill-climbing \cite{elsken2017simple}, MCTS \cite{negrinho2017deeparchitect}, SMBO \cite{mendoza2016towards,liu2017progressive}, and random search \cite{bergstra2012random}, and grid search \cite{zagoruyko2016wide}. Some methods even forewent the idea of independent architectures \cite{saxena2016convolutional}. There is much architecture-search work beyond image classification too, but that is outside our scope.

Even though some methods stand out due to their efficiency \cite{suganuma2017genetic,pham2018faster}, many approaches use large amounts of resources. Several recent papers reduced the compute cost through progressive-complexity search stages \cite{liu2017progressive}, hypernets \cite{brock2017smash}, accuracy prediction \cite{baker2017accelerating,klein2017learning,domhan2015speeding}, warm-starting and ensembling \cite{feurer2015efficient}, parallelization, reward shaping and early stopping \cite{zhong2017practical} or Net2Net transformations \cite{cai2017reinforcement}. Most of these methods could in principle be applied to evolution too, but this is beyond the scope of this paper.

A popular approach to evolution has been through \textit{generational} algorithms, \eg NEAT \cite{stanley2002evolving}. All models in the population must finish training before the next generation is computed. Generational evolution becomes inefficient in a distributed environment where a different machine is used to train each model: machines that train faster models finish earlier and must wait idle until all machines are ready. Real-time algorithms address this issue, \eg rtNEAT \cite{stanley2005real} and tournament selection \cite{goldberg1991comparative}. Unlike the generational algorithms, however, these discard models according to their performance or do not discard them at all, resulting in models that remain alive in the population for a long time---even for the whole experiment. We will present evidence that the finite lifetimes of aging evolution can give better results than direct tournament selection, while retaining its efficiency.

An existing paper \cite{hornby2006alps} uses a concept of \textit{age} but in a very different way than we do. In that paper, age is assigned to genes to divide a constant-size population into groups called \textit{age-layers}. Each layer contains individuals with genes of similar ages. Only after the genes have survived a certain \textit{age-gap}, they can make it to the next layer. The goal is to restrict competition (the newly introduced genes cannot be immediately out-competed by highly-selected older ones). Their algorithm requires the introduction of two additional meta-parameters (size of the age-gap and number of age-layers). In contrast, in our algorithm, an \textit{age} is assigned to the individuals (not the genes) and is only used to track which is the oldest individual in the population. This permits removing such oldest individual at each cycle (keeping a constant population size). Our approach, therefore, is in line with our goal of keeping the method as simple as possible. In particular, our method remains similar to nature (where the young are less likely to die than the very old) and it requires no additional meta-parameters.

\section{Methods}

This section contains a readable description of the methods. The Methods~Details section gives additional information.

\subsection{Search Space}

\begin{figure}[ht]
\centering
\raisebox{-0.5\height}{\includegraphics[height=0.55\linewidth]{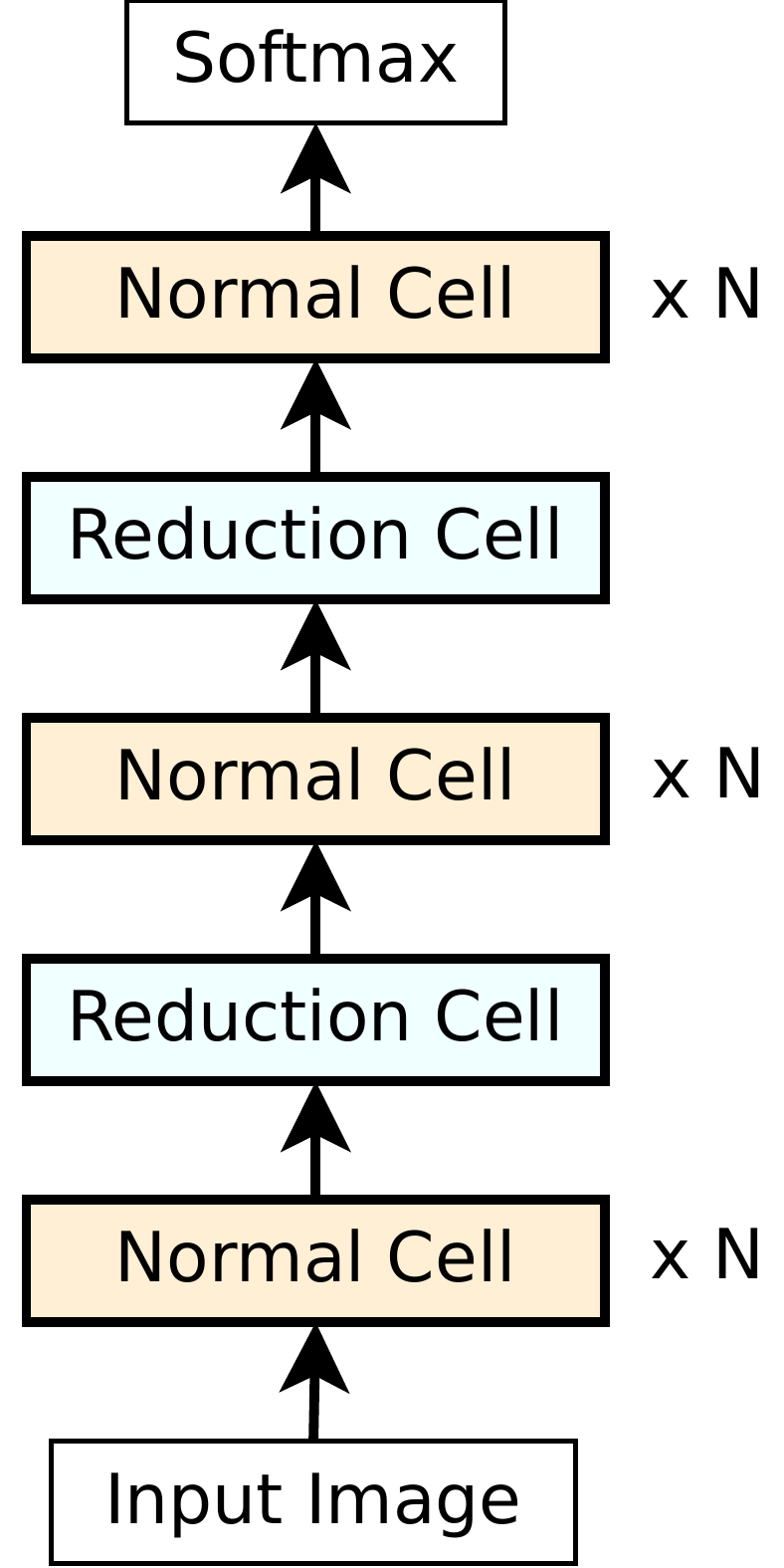}}
\hspace{0.02\linewidth}
\raisebox{-0.5\height}{\includegraphics[height=0.4\linewidth]{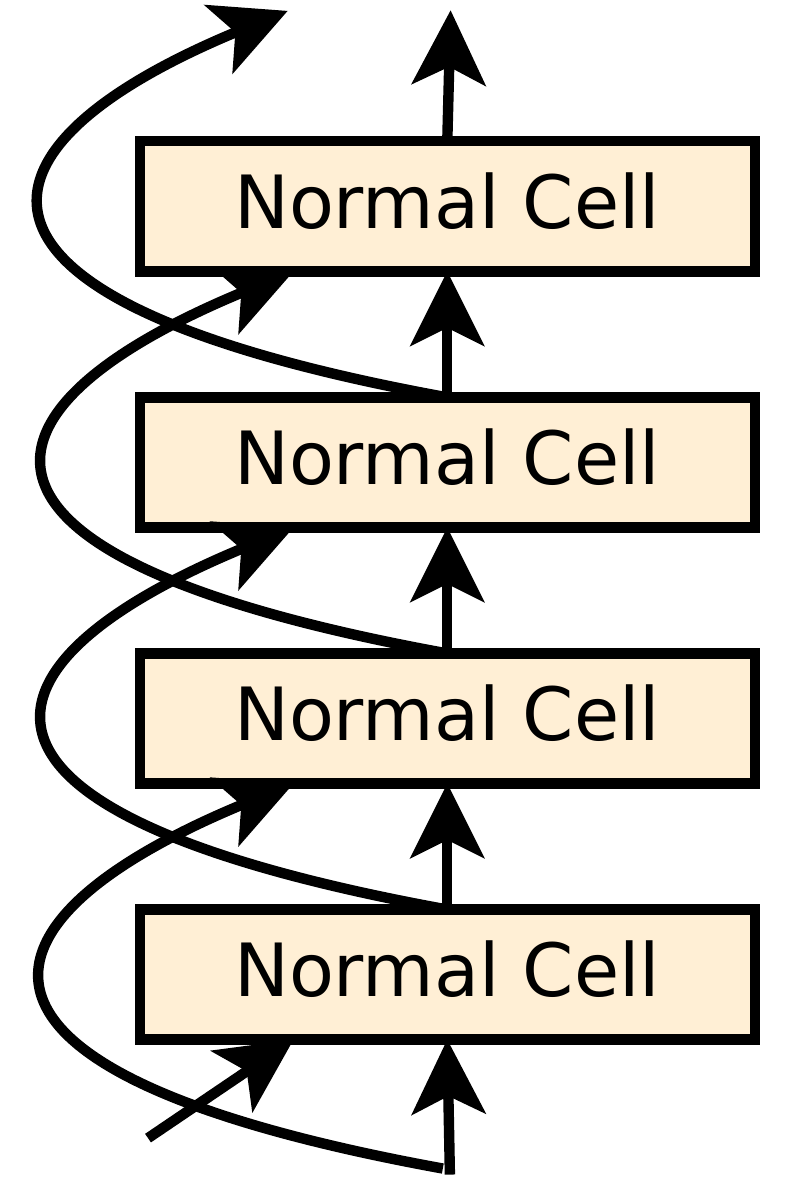}}
\hspace{0.05\linewidth}
\raisebox{-0.5\height}{\includegraphics[height=0.55\linewidth]{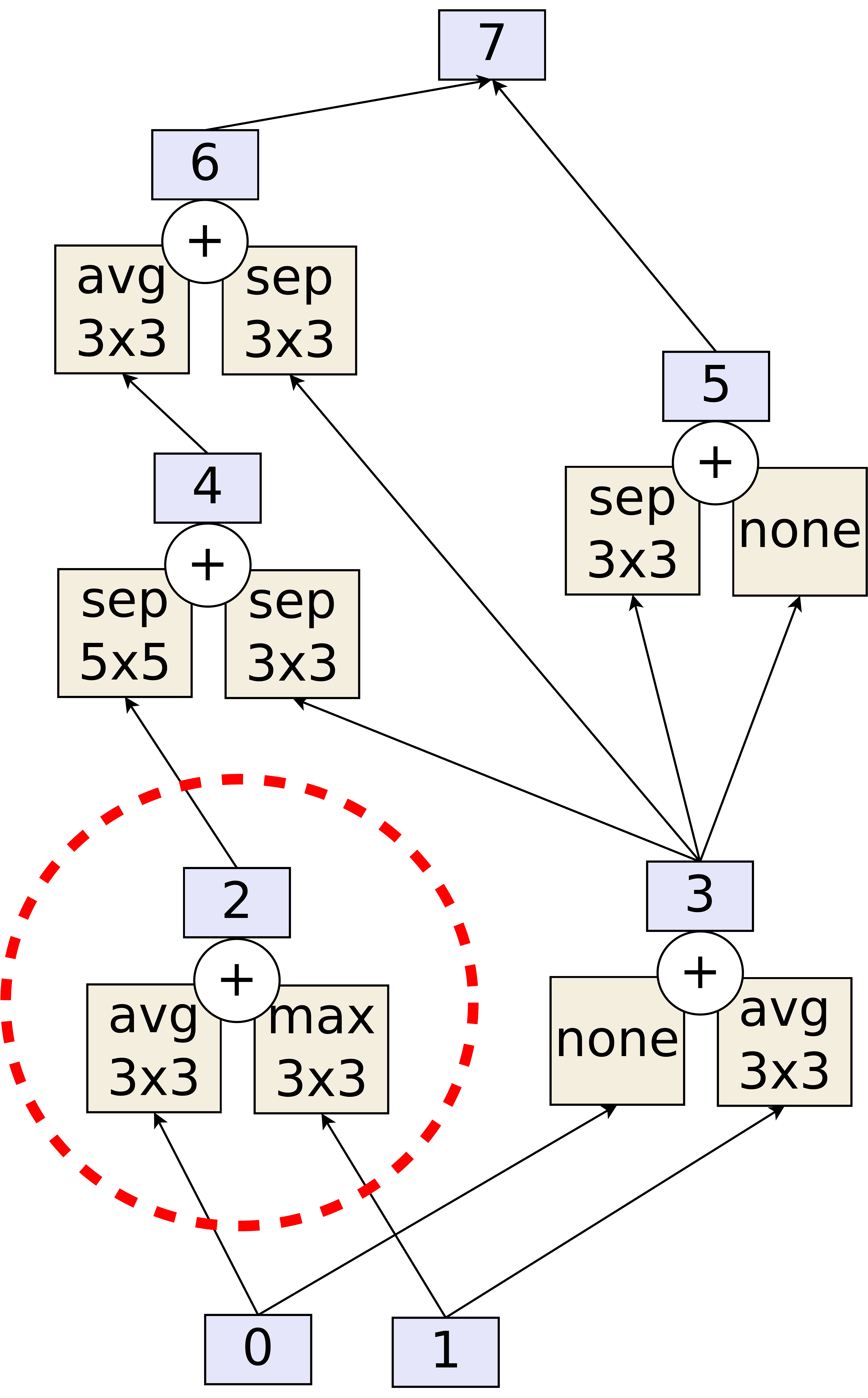}}
\caption{NASNet Search Space \cite{zoph2017learning}. LEFT: the full outer structure (omitting skip inputs for clarity). MIDDLE: detailed view with the skip inputs. RIGHT: cell example. Dotted line demarcates a pairwise combination.}
\label{search_space_fig}
\end{figure}

All experiments use the \textit{NASNet search space} \cite{zoph2017learning}. This is a space of image classifiers, all of which have the fixed outer structure indicated in Figure~\ref{search_space_fig} (left): a feed-forward stack of Inception-like modules called \textit{cells}. Each cell receives a \textit{direct input} from the previous cell (as depicted) and a \textit{skip input} from the cell before it (Figure~\ref{search_space_fig}, middle). The cells in the stack are of two types: the \textit{normal cell} and the \textit{reduction cell}. All normal cells are constrained to have the same architecture, as are reduction cells, but the architecture of the normal cells is independent of that of the reduction cells. Other than this, the only difference between them is that every application of the reduction cell is followed by a stride of 2 that reduces the image size, whereas normal cells preserve the image size. As can be seen in the figure, normal cells are arranged in three stacks of N cells. The goal of the architecture-search process is to discover the architectures of the normal and reduction cells.

As depicted in Figure~\ref{search_space_fig} (middle and right), each cell has two input activation tensors and one output. The very first cell takes two copies of the input image. After that, the inputs are the outputs of the previous two cells.

Both normal and reduction cells must conform to the following construction. The two cell input tensors are considered hidden states ``0'' and ``1''. More hidden states are then constructed through \textit{pairwise combinations}. A pairwise combination is depicted in Figure~\ref{search_space_fig} (right, inside dashed circle). It consists in applying an operation (or \textit{op}) to an existing hidden state, applying another op to another existing hidden state, and adding the results to produce a new hidden state. Ops belong to a fixed set of common convnet operations such as convolutions and pooling layers. Repeating hidden states or operations within a combination is permitted. In the cell example of Figure~\ref{search_space_fig} (right), the first pairwise combination applies a 3x3 average pool op to hidden state 0 and a 3x3 max pool op to hidden state 1, in order to produce hidden state 2. The next pairwise combination can now choose from hidden states 0, 1, and 2 to produce hidden state 3 (chose 0 and 1 in Figure~\ref{search_space_fig}), and so on. After exactly five pairwise combinations, any hidden states that remain unused (hidden states 5 and 6 in Figure~\ref{search_space_fig}) are concatenated to form the output of the cell (hidden state 7).

A given architecture is fully specified by the five pairwise combinations that make up the normal cell and the five that make up the reduction cell. Once the architecture is specified, the model still has two free parameters that can be used to alter its size (and its accuracy): the number of normal cells per stack (N) and the number of output filters of the convolution ops (F). N and F are determined manually.

\subsection{Evolutionary Algorithm}

The evolutionary method we used is summarized in Algorithm~\ref{aging_evol_alg}. It keeps a population of P trained models throughout the experiment. The population is initialized with models with random architectures (``{while $\left\vert{population}\right\vert$}'' in Algorithm~\ref{aging_evol_alg}). All architectures that conform to the search space described are possible and equally likely.

\begin{algorithm}
\caption{Aging Evolution}
\label{aging_evol_alg}
\begin{algorithmic}
\State $population \gets $ empty queue  \Comment The population.
\State $history \gets \varnothing$  \Comment Will contain all models.
\While{$\left\vert{population}\right\vert < P$}  \Comment Initialize population.
    \State $model.arch \gets \Call{RandomArchitecture}{ }$
    \State $model.accuracy \gets \Call{TrainAndEval}{model.arch}$
    \State add $model$ to right of $population$
    \State add $model$ to history
\EndWhile
\While{$\left\vert{history}\right\vert < C $}  \Comment Evolve for $C$ cycles.
    \State $sample \gets \varnothing$  \Comment Parent candidates.
    \While{$\left\vert{sample}\right\vert < S$}
        \State $candidate \gets$ random element from $population$
        \State \Comment The element stays in the $population$.
        \State add $candidate$ to $sample$
    \EndWhile
    \State $parent \gets$ highest-accuracy model in $sample$
    \State $child.arch \gets \Call{Mutate}{parent.arch}$
    \State $child.accuracy \gets \Call{TrainAndEval}{child.arch}$
    \State add $child$ to right of $population$
    \State add $child$ to $history$
    \State remove $dead$ from left of $population$  \Comment Oldest.
    \State discard $dead$
\EndWhile
\State \Return highest-accuracy model in $history$
\end{algorithmic}
\end{algorithm}

After this, evolution improves the initial population in cycles (``{while $\left\vert{history}\right\vert$}'' in Algorithm~\ref{aging_evol_alg}). At each cycle, it samples S random models from the population, each drawn uniformly at random with replacement. The model with the highest validation fitness within this sample is selected as the \textit{parent}. A new architecture, called the \textit{child}, is constructed from the parent by the application of a transformation called a \textit{mutation}. A mutation causes a simple and random modification of the architecture and is described in detail below. Once the child architecture is constructed, it is then trained, evaluated, and added to the population. This process is called tournament selection \cite{goldberg1991comparative}.

It is common in tournament selection to keep the population size fixed at the initial value P. This is often accomplished with an additional step within each cycle: discarding (or \textit{killing}) the worst model in the random S-sample. We will refer to this approach as \textit{non-aging evolution}. In contrast, in this paper we prefer a novel approach: killing the oldest model in the population---that is, removing from the population the model that was trained the earliest (``{remove \textit{dead} from left of \textit{pop}}'' in Algorithm~\ref{aging_evol_alg}). This favors the newer models in the population. We will refer to this approach as \textit{aging evolution}. In the context of architecture search, aging evolution allows us to explore the search space more, instead of zooming in on good models too early, as non-aging evolution would (see Discussion section for details).

In practice, this algorithm is parallelized by distributing the ``{while $\left\vert{history}\right\vert$}'' loop in Algorithm~\ref{aging_evol_alg} over multiple workers. A full implementation can be found online.\footnote{\url{https://colab.research.google.com/github/google-research/google-research/blob/master/evolution/regularized_evolution_algorithm/regularized_evolution.ipynb}} Intuitively, the mutations can be thought of as providing exploration, while the parent selection provides exploitation. The parameter $S$ controls the aggressiveness of the exploitation: $S=1$ reduces to a type of random search and $2 \le S \le P$ leads to evolution of varying greediness.

New models are constructed by applying a mutation to existing models, transforming their architectures in random ways. To navigate the NASNet search space described above, we use two main mutations that we call the \textit{hidden state mutation} and the \textit{op mutation}. A third mutation, the identity, is also possible. Only one of these mutations is applied in each cycle, choosing between them at random.

\begin{figure}[ht]
\centering
\includegraphics[width=0.67\linewidth]{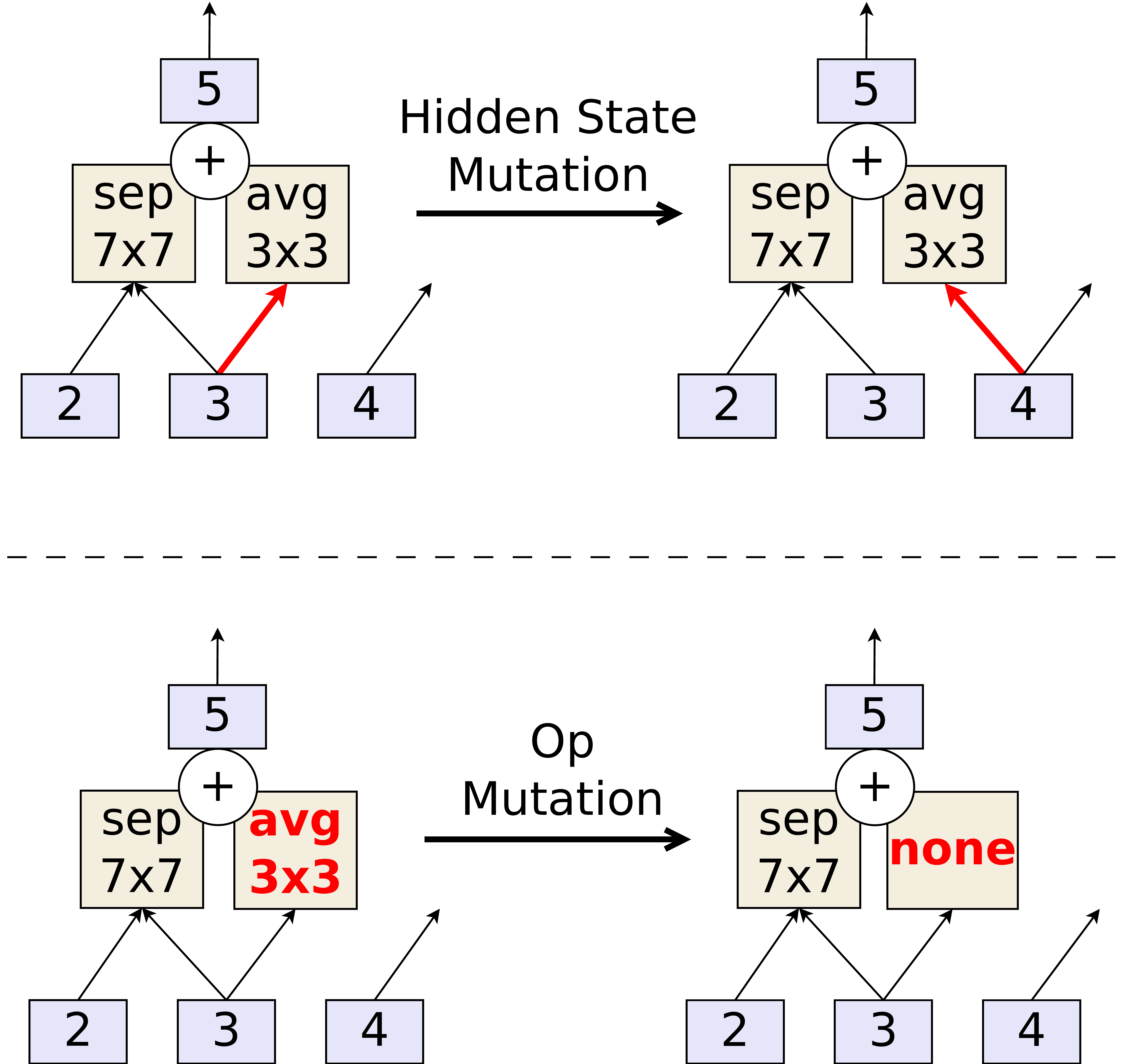}
\caption{Illustration of the two mutation types.}
\label{mutations_fig}
\end{figure}

The hidden state mutation consists of first making a random choice of whether to modify the normal cell or the reduction cell. Once a cell is chosen, the mutation picks one of the five pairwise combinations uniformly at random. Once the pairwise combination is picked, one of the two elements of the pair is chosen uniformly at random. The chosen element has one hidden state. This hidden state is now replaced with another hidden state from within the cell, subject to the constraint that no loops are formed (to keep the feed-forward nature of the convnet). Figure~\ref{mutations_fig} (top) shows an example.

The op mutation behaves like the hidden state mutation as far as choosing one of the two cells, one of the five pairwise combinations, and one of the two elements of the pair. Then it differs in that it modifies the op instead of the hidden state. It does this by replacing the existing op with a random choice from a fixed list of ops (see Methods~Details). Figure~\ref{mutations_fig} (bottom) shows an example.

\subsection{Baseline Algorithms}

Our main baseline is the application of RL to the same search space. RL was implemented using the algorithm and code in the baseline study \cite{zoph2017learning}. An LSTM controller outputs the architectures, constructing the pairwise combinations one at a time, and then gets a reward for each architecture by training and evaluating it. More detail can be found in the baseline study. We also compared against random search (RS). In our RS implementation, each model is constructed randomly so that all models in the search space are equally likely, as in the initial population in the evolutionary algorithm. In other words, the models in RS experiments are \mbox{\textit{not} constructed} by mutating existing models, so as to make new models independent from previous ones.

\subsection{Experimental Setup}

We ran controlled comparisons at scale, ensuring identical conditions for evolution, RL and random search (RS). In particular, all methods used \textit{the same} computer code for network construction, training and evaluation. Experiments always searched on the CIFAR-10 dataset \cite{krizhevsky2009learning}.

As in the baseline study, we first performed architecture search over small models (\ie small N and F) until 20k models were evaluated. After that, we used the \textit{model augmentation} trick \cite{zoph2017learning}: we took architectures discovered by the search (\eg the output of an evolutionary experiment) and turn them into a full-size, accurate models. To accomplish this, we enlarged the models by increasing N and F so the resulting model sizes would match the baselines, and we trained the enlarged models for a longer time on the CIFAR-10 or the ImageNet classification datasets \cite{krizhevsky2009learning,deng2009imagenet}. For ImageNet, a stem was added at the input of the model to reduce the image size, as shown in Figure~\ref{amoeba_a_arch} (left). This is the same procedure as in the baseline study. To produce the largest model (see last paragraph of Results section; not included in tables), we increased N and F until we ran out of memory. Actual values of N and F for all models are listed in the \mbox{Methods Details} section.

\section{Methods Details}

This section complements the Methods section with the details necessary to reproduce our experiments. Possible ops: none (identity); 3x3, 5x5 and 7x7 separable (sep.) convolutions (convs.); 3x3 average (avg.) pool; 3x3 max pool; 3x3 dilated (dil.) sep. conv.; 1x7 then 7x1 conv. Evolved with $P$=$100$, $S$=$25$. CIFAR-10 dataset \cite{krizhevsky2009learning} with 5k withheld examples for validation. Standard ImageNet dataset \cite{deng2009imagenet}, 1.2M 331x331 images and 1k classes; 50k examples withheld for validation; standard validation set used for testing. During the search phase, each model trained for 25 epochs; N=3/F=24, 1 GPU. Each experiment ran on 450 K40 GPUs for 20k models (approx.\ 7 days). To optimize evolution, we tried 5 configurations with P/S of: 100/2, 100/50, 20/20, 100/25, 64/16, best was 100/25. The probability of the identity mutation was fixed at the small, arbitrary value of 0.05 and was not tuned. Other mutation probabilities were uniform, as described in the Methods. To optimize RL, started with parameters already tuned in the baseline study and further optimized learning rate in 8 configurations: 0.00003, 0.00006, 0.00012, 0.0002, 0.0004, 0.0008, 0.0016, 0.0032; best was 0.0008. To avoid selection bias, plots do not include optimization runs, as was decided a priori. Best few (20) models were selected from each experiment and augmented to N=6/F=32, as in baseline study; batch 128, SGD with momentum rate 0.9, L2 weight decay $5 \times 10^{-4}$, initial lr 0.024 with cosine decay, 600 epochs, ScheduledDropPath to 0.7 prob; auxiliary softmax with half-weight of main softmax. For Table~\ref{evol_rl_semi_tab}, we used N/F of 6/32 and 6/36. For ImageNet table, N/F were 6/190 and 6/448 and standard training methods \cite{szegedy2017inception}: distributed sync SGD with 100 P100 GPUs; RMSProp optimizer with 0.9 decay and $\epsilon$=0.1, $4 \times 10^{-5}$ weight decay, 0.1 label smoothing, auxiliary softmax weighted by 0.4; dropout probability 0.5; ScheduledDropPath to 0.7 probability (as in baseline---note that this trick only contributes ~0.3\% top-1 ImageNet acc.); 0.001 initial lr, decaying every 2 epochs by 0.97. Largest model used N=6/F=448. F always refers to the number of filters of convolutions in the first stack; after each reduction cell, this number is doubled. Wherever applicable, we used the same conditions as the baseline study.

\section{Results}

\subsection{Comparison With RL and RS Baselines}

Currently, reinforcement learning (RL) is the predominant method for architecture search. In fact, today's state-of-the-art image classifiers have been obtained by architecture search with RL~\cite{zoph2017learning,liu2017progressive}. Here we seek to compare our evolutionary approach against their RL algorithm.  We performed large-scale side-by-side architecture-search experiments on CIFAR-10. We first optimized the hyper-parameters of the two approaches independently (details in Methods~Details section). Then we ran 5 repeats of each of the two algorithms---and also of random search (RS).

\begin{figure}[ht]
\centering
\includegraphics[width=0.67\linewidth]{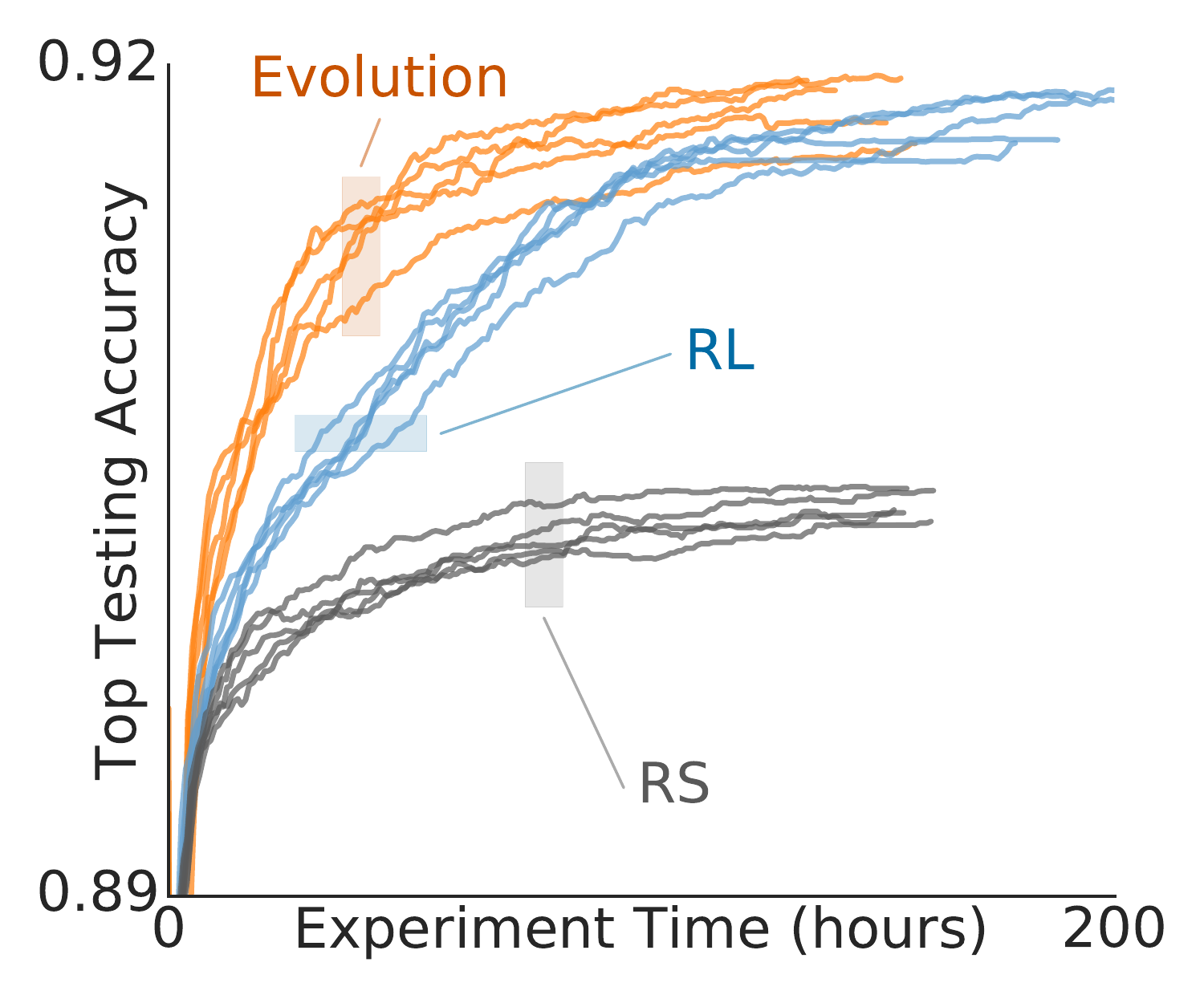}
\caption{Time-course of 5 identical large-scale experiments for each algorithm (evolution, RL, and RS), showing accuracy before augmentation on CIFAR-10. All experiments were stopped when 20k models were evaluated, as done in the baseline study. Note this plot does not show the compute cost of models, which was higher for the RL ones.}
\label{evol_rl_rs_progress_fig}
\end{figure}

Figure~\ref{evol_rl_rs_progress_fig} shows the model accuracy as the experiments progress, highlighting that evolution yielded more accurate models at the earlier stages, which could become important in a resource-constrained regime where the experiments may have to be stopped early (for example, when 450 GPUs for 7 days is too much). At the later stages, if we allow to run for the full 20k models (as in the baseline study), evolution produced models with similar accuracy. Both evolution and RL compared favorably against RS. It is important to note that the vertical axis of Figure~\ref{evol_rl_rs_progress_fig} does not present the compute cost of the models, only their accuracy. Next, we will consider their compute cost as well.

As in the baseline study, the architecture-search experiments above were performed over small models, to be able to train them quicker. We then used the \textit{model augmentation} trick \cite{zoph2017learning} by which we take an architecture discovered by the search (\eg the output of an evolutionary experiment) and turn it into a full-size, accurate model, as described in the Methods.

\begin{figure}[ht]
\centering
\includegraphics[width=0.67\linewidth]{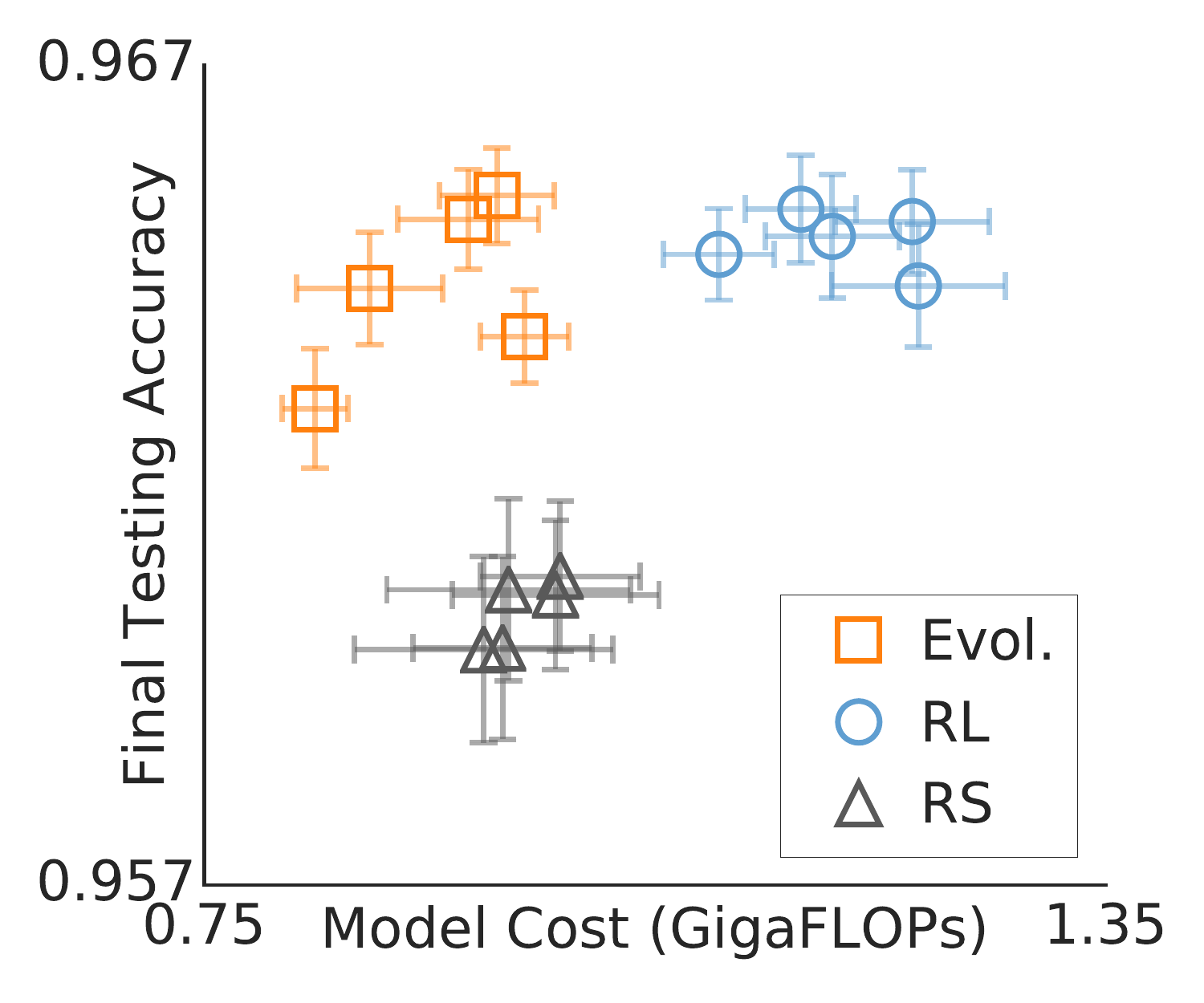}
\caption{Final augmented models from 5 identical architecture-search experiments for each algorithm, on CIFAR-10. Each marker corresponds to the top models from one experiment.}
\label{evol_rl_rs_aug_fig}
\end{figure}

\begin{figure*}[b!]
\centering
\includegraphics[height=0.3\linewidth]{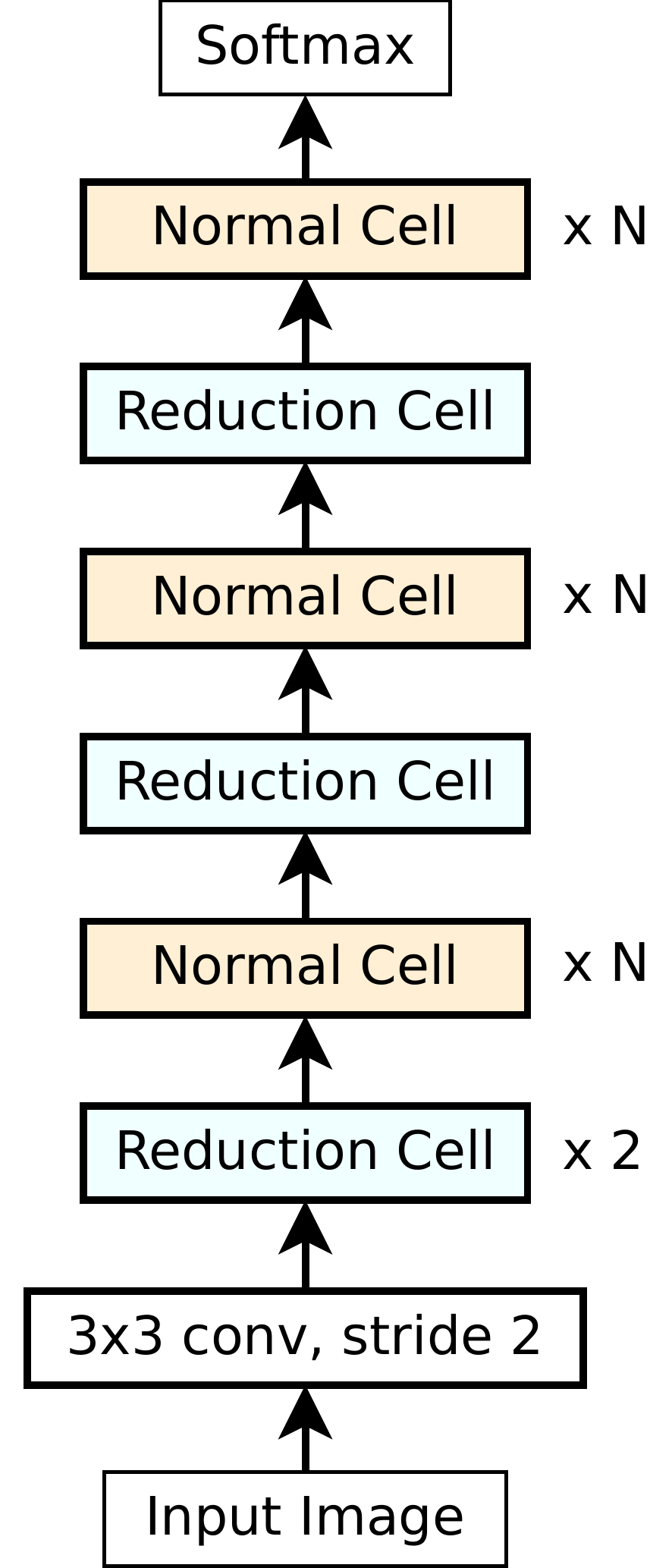}
\hspace{0.07\linewidth}
\includegraphics[height=0.3\linewidth]{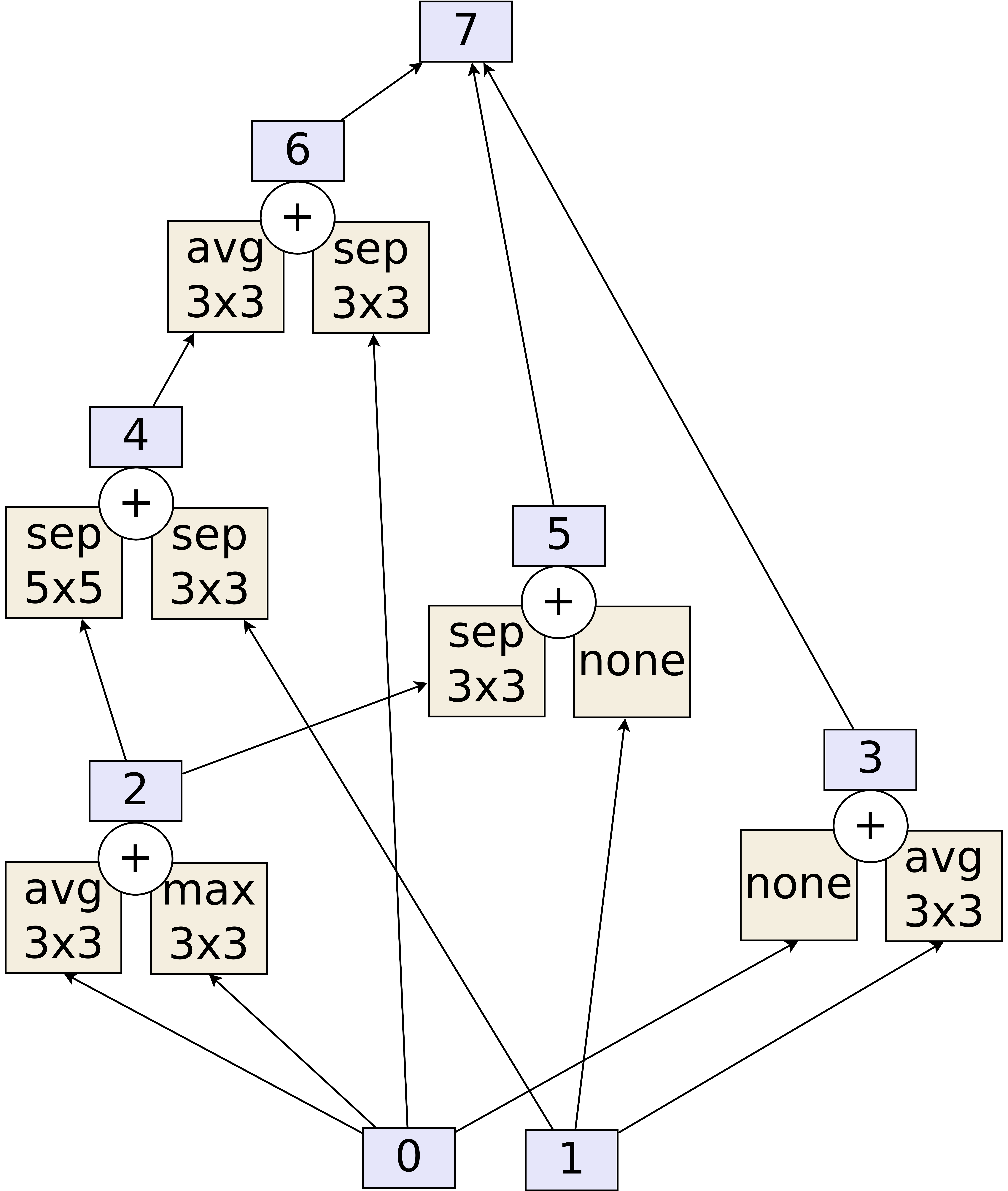}
\hspace{0.05\linewidth}
\includegraphics[height=0.3\linewidth]{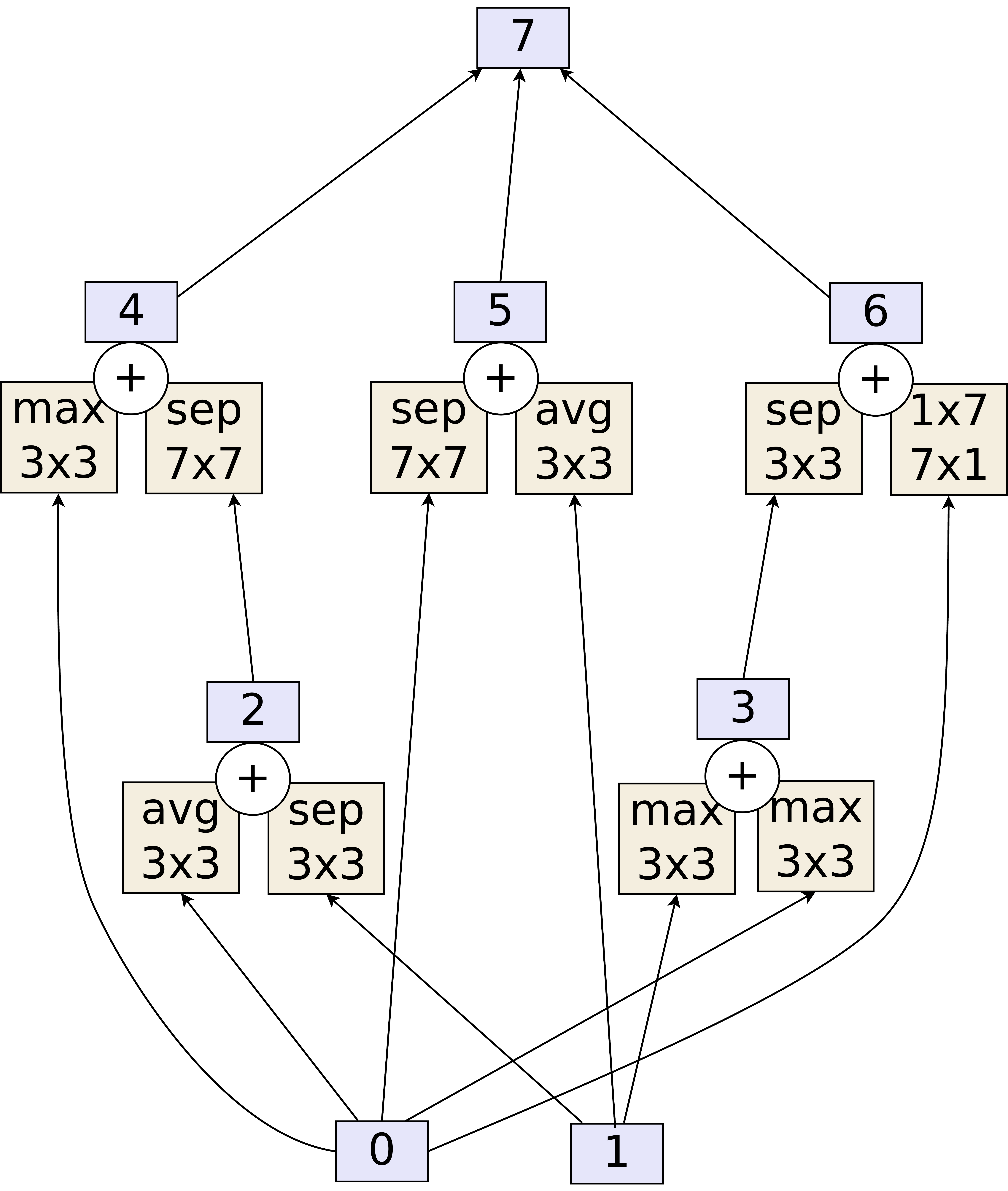}
\caption{\mbox{AmoebaNet-A} architecture. The overall model \cite{zoph2017learning} (LEFT) and the \mbox{AmoebaNet-A} normal cell (MIDDLE) and reduction cell (RIGHT).}
\label{amoeba_a_arch}
\end{figure*}

Figure~\ref{evol_rl_rs_aug_fig} compares the augmented top models from the three sets of experiments. It shows test accuracy and model compute cost. The latter is measured in FLOPs, by which we mean the total count of operations in the forward pass, so lower is better. Evolved architectures had higher accuracy (and similar FLOPs) than those obtained with RS, and lower FLOPs (and similar accuracy) than those obtained with RL. Number of parameters showed similar behavior to FLOPs. Therefore, evolution occupied the ideal relative position in this graph within the scope of our case study.

So far we have been comparing evolution with our reproduction of the experiments in the baseline study, but it is also informative to compare directly against the results reported by the baseline study. We select our evolved architecture with highest validation accuracy and call it \mbox{\textit{AmoebaNet-A}} (Figure~\ref{amoeba_a_arch}). Table~\ref{evol_rl_semi_tab} compares its test accuracy with the top model of the baseline study, NASNet-A. Such a comparison is not entirely controlled, as we have no way of ensuring the network training code was identical and that the same number of experiments were done to obtain the final model. The table summarizes the results of training \mbox{AmoebaNet-A} at sizes comparable to a NASNet-A version, showing that \mbox{AmoebaNet-A} is slightly more accurate (when matching model size) or considerably smaller (when matching accuracy). We did not train our model at larger sizes on CIFAR-10. Instead, we moved to ImageNet to do further comparisons in the next section.

\begin{table}[ht]
\caption{CIFAR-10 testing set results for \mbox{AmoebaNet-A}, compared to top model reported in the baseline study.}
\label{evol_rl_semi_tab}
\centering
\begin{tabular}{lcc}
\toprule
Model                       &  \# Params & Test Error (\%)   \\
\midrule
NASNet-A (baseline)    & 3.3 M   & $3.41$  \\
AmoebaNet-A (N=6, F=32)   & 2.6 M   & $3.40 \pm 0.08$  \\
AmoebaNet-A (N=6, F=36)   & 3.2 M   & $3.34 \pm 0.06$  \\
\bottomrule
\end{tabular}
\end{table}

\subsection{ImageNet Results}

Following the accepted standard, we compare our top model's classification accuracy on the popular ImageNet dataset against other top models from the literature. Again, we use \mbox{AmoebaNet-A}, the model with the highest validation accuracy on CIFAR-10 among our evolution experiments. We highlight that the model was evolved on CIFAR-10 and then transferred to ImageNet, so the evolved \textit{architecture} cannot have overfit the ImageNet dataset. When re-trained on ImageNet, \mbox{AmoebaNet-A} performs comparably to the baseline for the same number of parameters (Table~\ref{results_imagenet_tab}, model with F=190).

\begin{table*}[ht]
\caption{ImageNet classification results for \mbox{AmoebaNet-A} compared to hand-designs (top rows) and other automated methods (middle rows). The evolved AmoebaNet-A architecture (bottom rows) reaches the current state of the art (SOTA) at similar model sizes and sets a new SOTA at a larger size. All evolution-based approaches are marked with a \textsuperscript{\textasteriskcentered}. We omitted Squeeze-and-Excite-Net because it was not benchmarked on the same ImageNet dataset version.}
\label{results_imagenet_tab}
\centering
\begin{tabular}{lccc}
\toprule
Model                       &  \# Parameters & \# Multiply-Adds & Top-1 / Top-5 Accuracy (\%)   \\
\midrule
Incep-ResNet V2 \cite{szegedy2017inception}   & 55.8M   & 13.2B      & 80.4 / 95.3 \\
ResNeXt-101 \cite{xie2017aggregated}    & 83.6M   & 31.5B      & 80.9 / 95.6 \\  
PolyNet \cite{zhang2017polynet}         & 92.0M   & 34.7B      & 81.3 / 95.8 \\
Dual-Path-Net-131 \cite{chen2017dual}      & 79.5M   & 32.0B      & 81.5 / 95.8 \\
\midrule 
GeNet-2 \cite{xie2017genetic}\textsuperscript{\textasteriskcentered}    & 156M    & --        & 72.1 / 90.4 \\
Block-QNN-B \cite{zhong2017practical}\textsuperscript{\textasteriskcentered} & -- & -- & 75.7 / 92.6  \\  
Hierarchical \cite{liu2017hierarchical}\textsuperscript{\textasteriskcentered} & 64M & -- & 79.7 / 94.8 \\  
NASNet-A \cite{zoph2017learning}  & 88.9M   & 23.8B      & 82.7 / 96.2  \\  
PNASNet-5 \cite{liu2017progressive}  & 86.1M   & 25.0B      & 82.9 / 96.2  \\  
\midrule
AmoebaNet-A (N=6, F=190)\textsuperscript{\textasteriskcentered} & 86.7M  &  23.1B      & 82.8 / 96.1 \\
AmoebaNet-A (N=6, F=448)\textsuperscript{\textasteriskcentered} & 469M  &  104B      & 83.9 / 96.6 \\
\bottomrule
\end{tabular}
\end{table*}

Finally, we focused on \mbox{AmoebaNet-A} exclusively and enlarged it, setting a new state-of-the-art accuracy on ImageNet of 83.9\%/96.6\% top-1/5 accuracy with 469M parameters (Table~\ref{results_imagenet_tab}, model with F=448). Such high parameter counts may be beneficial in training other models too but we have not managed to do this yet.

\section{Discussion}

This section will suggest directions for future work, which we will motivate by speculating about the evolutionary process and by summarizing additional minor results. The details of these minor results have been relegated to the supplements, as they are not necessary to understand or reproduce our main results above.

\begin{figure}[ht]
\centering
\includegraphics[width=0.98\linewidth]{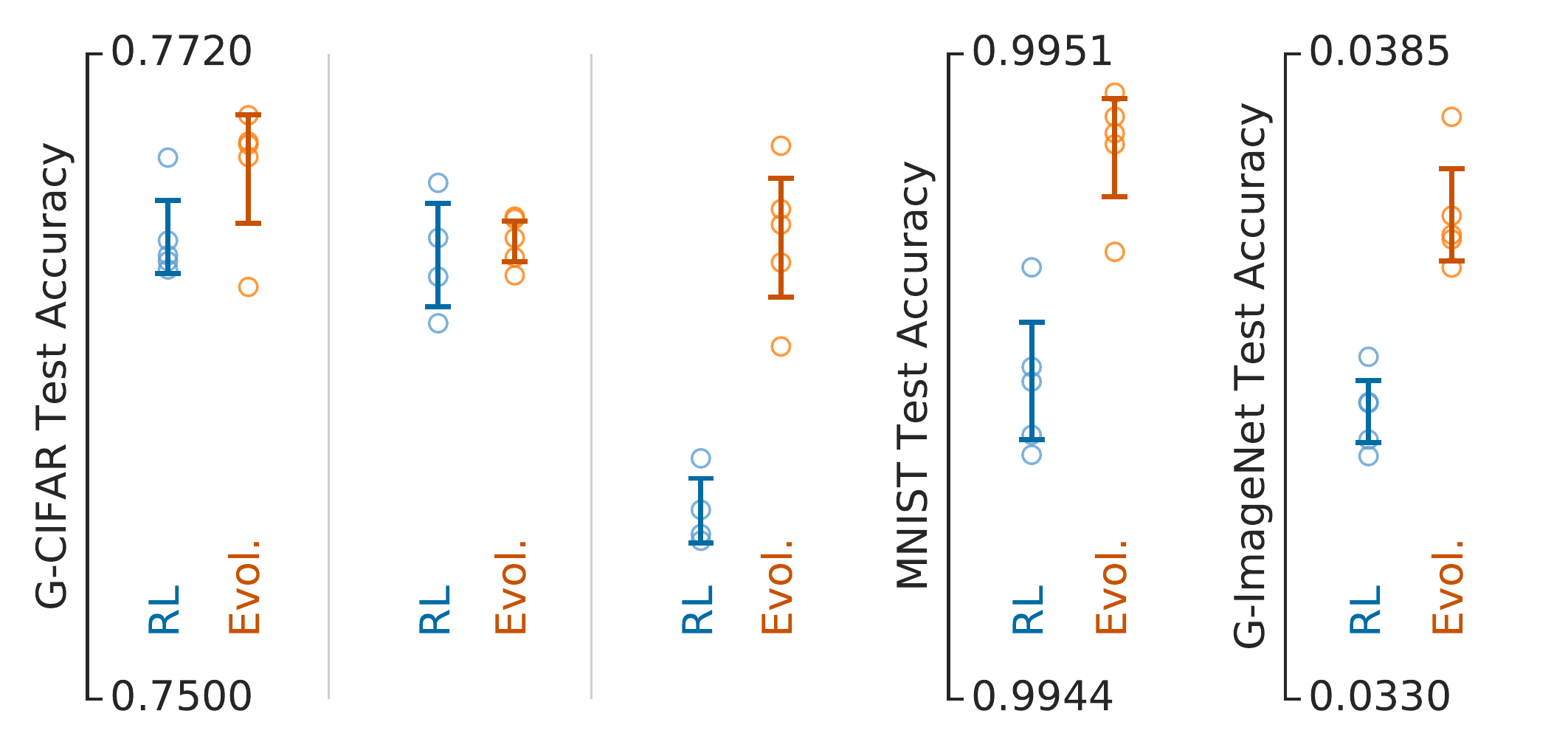}
\includegraphics[width=0.49\linewidth]{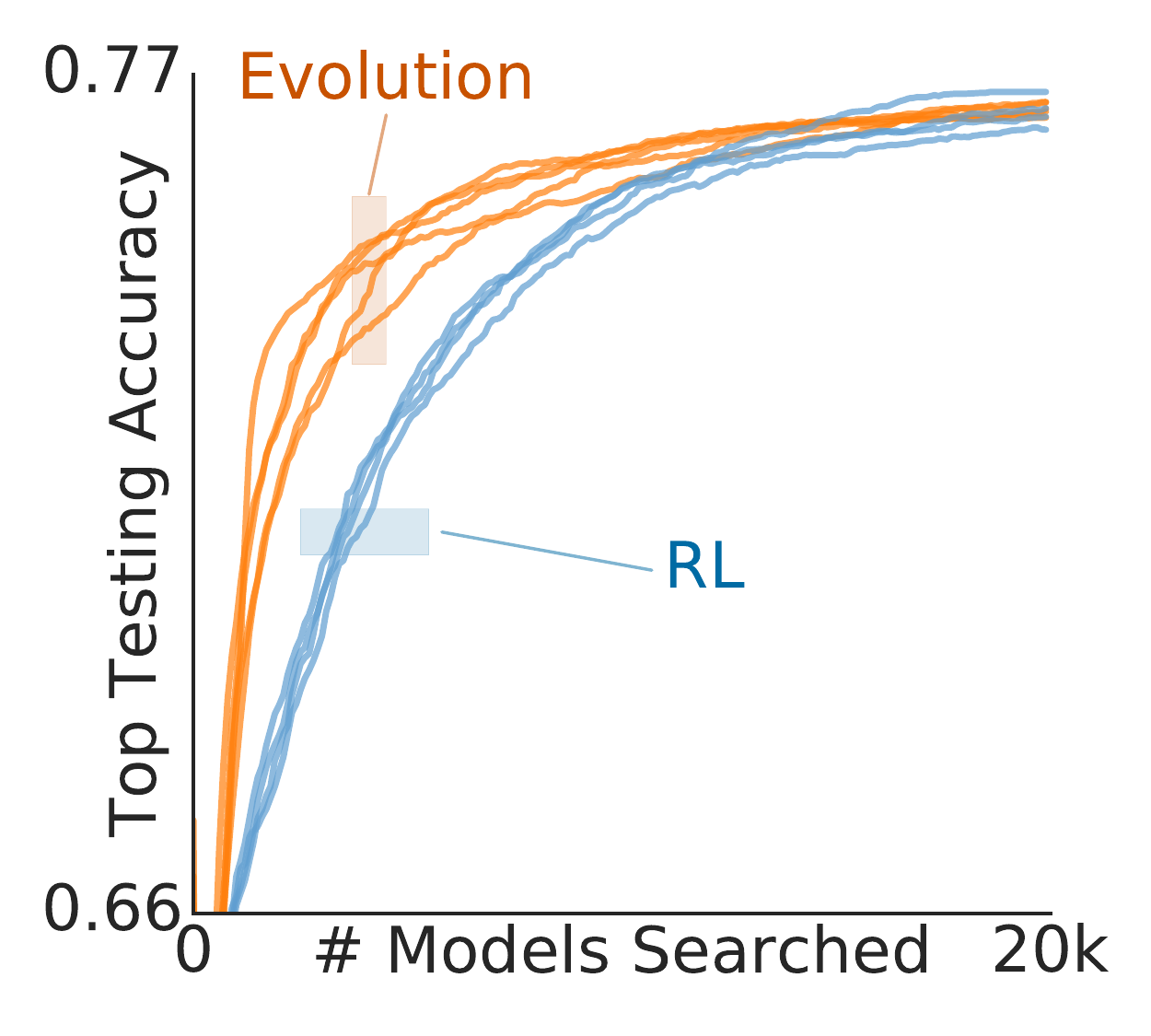}
\includegraphics[width=0.49\linewidth]{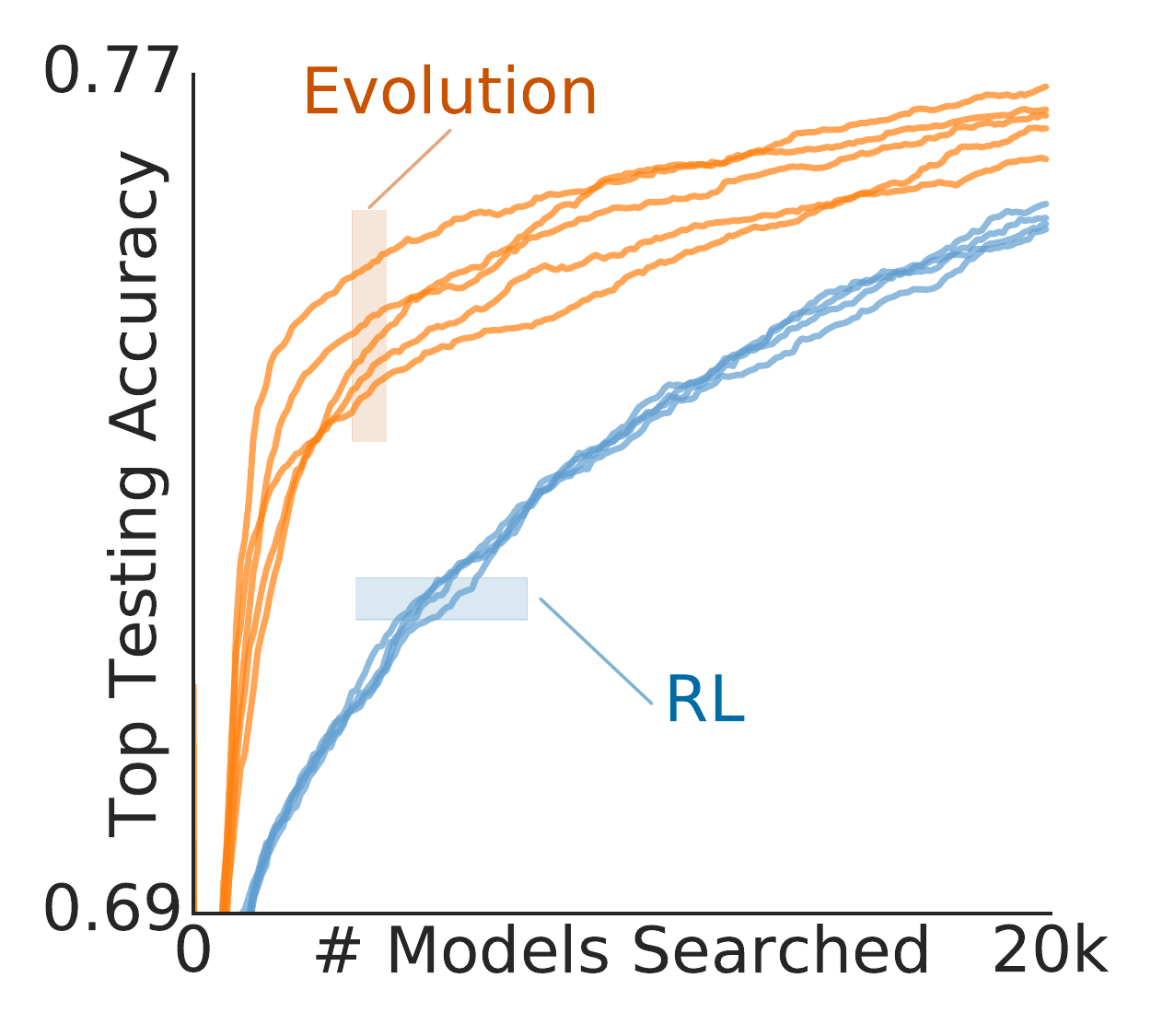}
\caption{TOP: Comparison of the final model accuracy in five different contexts, from left to right: G-CIFAR/SP-I, G-CIFAR/SP-II, G-CIFAR/SP-III, MNIST/SP-I and G-ImageNet/SP-I. Each circle marks the top test accuracy at the end of one experiment. BOTTOM: Search progress of the experiments in the case of G-CIFAR/SP-II (LEFT, best for RL) and G-CIFAR/SP-III (RIGHT, best for evolution).}
\label{evol_vs_rl_small_fig}
\end{figure}

\textbf{Scope of results.} Some of our findings may be restricted to the search spaces and datasets we used. A natural direction for future work is to extend the controlled comparison to more search spaces, datasets, and tasks, to verify generality, or to more algorithms. Supplement~A presents preliminary results, performing evolutionary and RL searches over three search spaces (SP-I: same as in the Results section; SP-II: like SP-I but with more possible ops; SP-III: like SP-II but with more pairwise combinations) and three datasets (gray-scale CIFAR-10, MNIST, and gray-scale ImageNet), at a small-compute scale (on CPU, $F$=$8$, $N$=$1$). Evolution reached equal or better accuracy in all cases (Figure~\ref{evol_vs_rl_small_fig}, top).

\textbf{Algorithm speed.} In our comparison study, Figure~\ref{evol_rl_rs_progress_fig} suggested that both RL and evolution are approaching a common accuracy asymptote. That raises the question of which algorithm gets there faster. The plots indicate that evolution reaches half-maximum accuracy in roughly half the time. We abstain, nevertheless, from further quantifying this effect since it depends strongly on how speed is measured (the number of models necessary to reach accuracy $a$ depends on $a$; the natural choice of $a=a_{max} / 2$ may be too low to be informative; \etc). Algorithm speed may be more important when exploring larger spaces, where reaching the optimum can require more compute than is available. We saw an example of this in the SP-III space, where evolution stood out (Figure~\ref{evol_vs_rl_small_fig}, bottom-right). Therefore, future work could explore evolving on even larger spaces.

\textbf{Model speed.} The speed of individual models produced is also relevant. Figure~\ref{evol_rl_rs_aug_fig} demonstrated that evolved models are faster (lower FLOPs). We speculate that asynchronous evolution may be reducing the FLOPs because it is indirectly optimizing for speed even when training for a fixed number of epochs: fast models may do well because they ``reproduce'' quickly even if they initially lack the higher accuracy of their slower peers. Verifying this speculation could be the subject of future work. As mentioned in the Related Work section, in this work we only considered asynchronous algorithms (as opposed to generational evolutionary methods) to ensure high resource utilization. Future work may explore how asynchronous and generational algorithms compare with regard to model accuracy.

\textbf{Benefits of aging evolution.} Aging evolution seemed advantageous in additional small-compute-scale experiments, shown in Figure~\ref{aging_vs_non_aging_fig} and presented in more detail in Supplement~B. These were carried out on CPU instead of GPU, and used a gray-scale version of CIFAR-10, to reduce compute requirements. In the supplement, we also show that these results tend to hold when varying the dataset or the search space.

\begin{figure}[ht]
\centering
\includegraphics[width=0.67\linewidth]{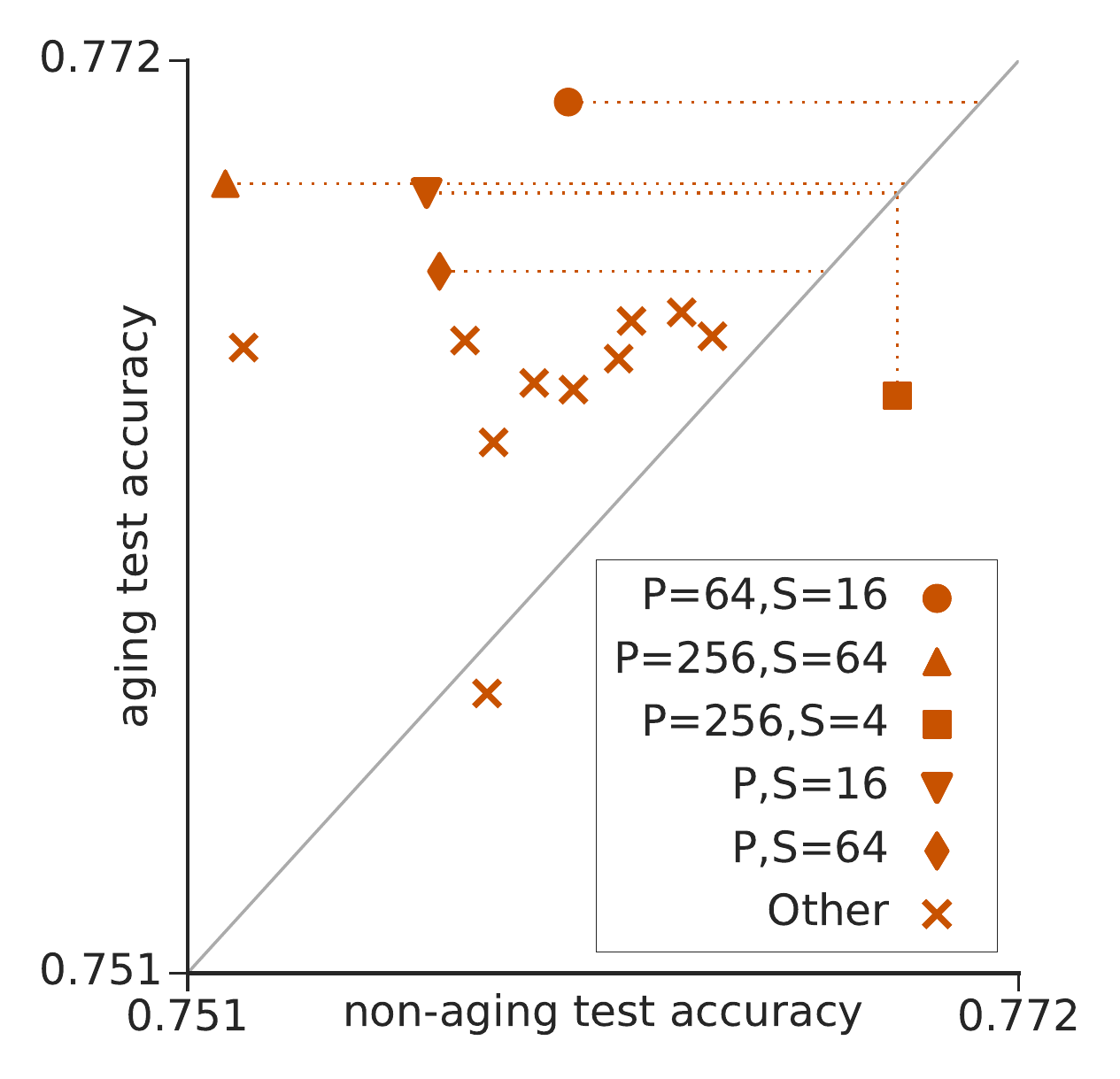}
\caption{Small-compute-scale comparison between our aging tournament selection variant and the non-aging variant, for different population sizes (P) and sample sizes (S), showing that aging tends to be beneficial (most markers are above the $y=x$ line).}
\label{aging_vs_non_aging_fig}
\end{figure}

\textbf{Understanding aging evolution and regularization.} We can speculate that aging may help navigate the training noise in evolutionary experiments, as follows. Noisy training means that models may sometimes reach high accuracy just by luck. In non-aging evolution (NAE, \ie standard tournament selection), such lucky models may remain in the population for a long time---even for the whole experiment. One lucky model, therefore, can produce many children, causing the algorithm to focus on it, reducing exploration. Under aging evolution (AE), on the other hand, all models have a short lifespan, so the population is wholly renewed frequently, leading to more diversity and more exploration. In addition, another effect may be in play, which we describe next. In AE, because models die quickly, the only way an architecture can remain in the population for a long time is by being passed down from parent to child through the generations. Each time an architecture is inherited it must be re-trained. If it produces an inaccurate model when re-trained, that model is not selected by evolution and the architecture disappears from the population. The only way for an architecture to remain in the population for a long time is to re-train well repeatedly. In other words, AE can only improve a population through the inheritance of architectures that re-train well. (In contrast, NAE can improve a population by accumulating architectures/models that were lucky when they trained the first time). That is, AE is forced to pay attention to \textit{architectures} rather than \textit{models}. In other words, the addition of aging involves introducing additional information to the evolutionary process: architectures should re-train well. This additional information prevents overfitting to the training noise, which makes it a form of \textit{regularization} in the broader mathematical sense\footnote{\url{https://en.wikipedia.org/wiki/Regularization_(mathematics)}}. Regardless of the exact mechanism, in Supplement~C we perform experiments to verify the plausibility of the conjecture that aging helps navigate noise. There we construct a toy search space where the only difficulty is a noisy evaluation. If our conjecture is true, AE should be better in that toy space too. We found this to be the case. We leave further verification of the conjecture to future work, noting that theoretical results may prove useful here.

\textbf{Simplicity of aging evolution.} A desirable feature of evolutionary algorithms is their simplicity. By design, the application of a mutation causes a random change. The process of constructing new architectures, therefore, is entirely random. What makes evolution different from random search is that only the good models are selected to be mutated. This selection tends to improve the population over time. In this sense, evolution is simply ``random search plus selection''. In outline, the process can be described briefly: ``keep a population of N models and proceed in cycles: at each cycle, copy-mutate the best of S random models and kill the oldest in the population''. Implementation-wise, we believe the methods of this paper are sufficient for a reader to understand evolution. The sophisticated nature of the RL alternative introduces complexity in its implementation: it requires back-propagation and poses challenges to parallelization \cite{salimans2017evolution}. Even different implementations of the same algorithm have been shown to produce different results \cite{henderson2017deep}. Finally, evolution is also simple in that it has few meta-parameters, most of which do not need tuning \cite{real2017large}. In our study, we only adjusted 2 meta-parameters and only through a handful of attempts (see Methods~Details section). In contrast, note that the RL baseline requires training an agent/controller which is often itself a neural network with many weights (such as an LSTM), and its optimization has more meta-parameters to adjust: learning rate schedule, greediness, batching, replay buffer, \etc. (These meta-parameters are all in addition to the weights and training parameters of the image classifiers being searched, which are present in both approaches.) It is possible that through careful tuning, RL could be made to produce even better models than evolution, but such tuning would likely involve running many experiments, making it more costly. Evolution did not require much tuning, as described. It is also possible that random search would produce equally good models if run for a very long time, which would be very costly.

\textbf{Interpreting architecture search.} Another important direction for future work is that of analyzing architecture-search experiments (regardless of the algorithm used) to try to discover new neural network design patterns. Anecdotally, for example, we found that architectures with high output vertex fan-in (number of edges into the output vertex) tend to be favored in all our experiments. In fact, the models in the final evolved populations have a mean fan-in value that is 3 standard deviations above what would be expected from randomly generated models. We verified this pattern by training various models with different fan-in values and the results confirm that accuracy increases with fan-in, as had been found in ResNeXt \cite{xie2017aggregated}. Discovering broader patterns may require designing search spaces specifically for this purpose.

\textbf{Additional AmoebaNets.} Using variants of the evolutionary process described, we obtained three additional models, which we named \mbox{\textit{AmoebaNet-B}}, \mbox{\textit{AmoebaNet-C}}, and \mbox{\textit{AmoebaNet-D}}. We describe these models and the process that led to them in detail in Supplement~D, but we summarize here. \mbox{AmoebaNet-B} was obtained through through platform-aware architecture search over a larger version of the NASNet space.  \mbox{AmoebaNet-C} is simply a model that showed promise early on in the above experiments by reaching high accuracy with relatively few parameters; we mention it here for completeness, as it has been referenced in other work\cite{cubuk2018autoaugment}. \mbox{AmoebaNet-D} was obtained by manually extrapolating the evolutionary process and optimizing the resulting architecture for training speed. It is very efficient: \mbox{AmoebaNet-D} won the Stanford DAWNBench competition for lowest training cost on ImageNet\cite{coleman2018analysis}.

\section{Conclusion}

This paper used an evolutionary algorithm to discover image classifier architectures. Our contributions are the following:
\begin{itemize}
    \item We proposed \textit{aging evolution}, a variant of tournament selection by which genotypes die according to their age, favoring the young. This improved upon standard tournament selection while still allowing for efficiency at scale through asynchronous population updating. We open-sourced the code.\footnote{\url{https://colab.research.google.com/github/google-research/google-research/blob/master/evolution/regularized_evolution_algorithm/regularized_evolution.ipynb}} We also implemented simple mutations that permit the application of evolution to the popular NASNet search space.
    \item We presented the first controlled comparison of algorithms for image classifier architecture search in a case study of evolution, RL and random search. We showed that evolution had somewhat faster search speed and stood out in the regime of scarcer resources / early stopping. Evolution also matched RL in final model quality, employing a simpler method.
    \item We evolved \mbox{AmoebaNet-A} (Figure~\ref{amoeba_a_arch}), a competitive image classifier. On ImageNet, it is the first evolved model to surpass hand-designs. Matching size, \mbox{AmoebaNet-A} has comparable accuracy to top image-classifiers discovered with other architecture-search methods. At large size, it sets a new state-of-the-art accuracy. We open-sourced code and checkpoint.\footnotemark{}.
\end{itemize}

\section{Acknowledgments}
We wish to thank Megan Kacholia, Vincent Vanhoucke, Xiaoqiang Zheng and especially Jeff Dean for their support and valuable input; Chris Ying for his work helping tune AmoebaNet models and for his help with specialized hardware, Barret Zoph and Vijay Vasudevan for help with the code and experiments used in their paper \cite{zoph2017learning}, as well as Jiquan Ngiam, Jacques Pienaar, Arno Eigenwillig, Jianwei Xie, Derek Murray, Gabriel Bender, Golnaz Ghiasi, Saurabh Saxena and Jie Tan for other coding contributions; Jacques Pienaar, Luke Metz, Chris Ying and Andrew Selle for manuscript comments, all the above and Patrick Nguyen, Samy Bengio, Geoffrey Hinton, Risto Miikkulainen, Jeff Clune, Kenneth Stanley, Yifeng Lu, David Dohan, David So, David Ha, Vishy Tirumalashetty, Yoram Singer, and Ruoming Pang for helpful discussions; and the larger Google Brain team.

\footnotetext{\url{https://tfhub.dev/google/imagenet/amoebanet_a_n18_f448/classification/1}}

\bibliographystyle{abbrv}
\bibliography{real2018_regularized_evolution}

\begin{thebibliography}{10}

\bibitem{angeline1994evolutionary}
P.~J. Angeline, G.~M. Saunders, and J.~B. Pollack.
\newblock An evolutionary algorithm that constructs recurrent neural networks.
\newblock {\em IEEE transactions on Neural Networks}, 1994.

\bibitem{baker2016designing}
B.~Baker, O.~Gupta, N.~Naik, and R.~Raskar.
\newblock Designing neural network architectures using reinforcement learning.
\newblock In {\em ICLR}, 2017.

\bibitem{baker2017accelerating}
B.~Baker, O.~Gupta, R.~Raskar, and N.~Naik.
\newblock Accelerating neural architecture search using performance prediction.
\newblock {\em ICLR Workshop}, 2017.

\bibitem{bergstra2012random}
J.~Bergstra and Y.~Bengio.
\newblock Random search for hyper-parameter optimization.
\newblock {\em JMLR}, 2012.

\bibitem{brock2017smash}
A.~Brock, T.~Lim, J.~M. Ritchie, and N.~Weston.
\newblock Smash: one-shot model architecture search through hypernetworks.
\newblock In {\em ICLR}, 2018.

\bibitem{cai2017reinforcement}
H.~Cai, T.~Chen, W.~Zhang, Y.~Yu, and J.~Wang.
\newblock Efficient architecture search by network transformation.
\newblock In {\em AAAI}, 2018.

\bibitem{chen2017dual}
Y.~Chen, J.~Li, H.~Xiao, X.~Jin, S.~Yan, and J.~Feng.
\newblock Dual path networks.
\newblock In {\em NIPS}, 2017.

\bibitem{ciregan2012multi}
D.~Ciregan, U.~Meier, and J.~Schmidhuber.
\newblock Multi-column deep neural networks for image classification.
\newblock In {\em CVPR}, 2012.

\bibitem{coleman2018analysis}
C.~Coleman, D.~Kang, D.~Narayanan, L.~Nardi, T.~Zhao, J.~Zhang, P.~Bailis,
  K.~Olukotun, C.~Re, and M.~Zaharia.
\newblock Analysis of dawnbench, a time-to-accuracy machine learning
  performance benchmark.
\newblock {\em arXiv preprint arXiv:1806.01427}, 2018.

\bibitem{cortes2016adanet}
C.~Cortes, X.~Gonzalvo, V.~Kuznetsov, M.~Mohri, and S.~Yang.
\newblock Adanet: Adaptive structural learning of artificial neural networks.
\newblock In {\em ICML}, 2017.

\bibitem{cubuk2018autoaugment}
E.~D. Cubuk, B.~Zoph, D.~Mane, V.~Vasudevan, and Q.~V. Le.
\newblock Autoaugment: Learning augmentation policies from data.
\newblock {\em arXiv}, 2018.

\bibitem{deng2009imagenet}
J.~Deng, W.~Dong, R.~Socher, L.-J. Li, K.~Li, and L.~Fei-Fei.
\newblock Imagenet: A large-scale hierarchical image database.
\newblock In {\em CVPR}, 2009.

\bibitem{domhan2015speeding}
T.~Domhan, J.~T. Springenberg, and F.~Hutter.
\newblock Speeding up automatic hyperparameter optimization of deep neural
  networks by extrapolation of learning curves.
\newblock In {\em IJCAI}, 2017.

\bibitem{elsken2017simple}
T.~Elsken, J.-H. Metzen, and F.~Hutter.
\newblock Simple and efficient architecture search for convolutional neural
  networks.
\newblock {\em ICLR Workshop}, 2017.

\bibitem{elsken2018neural}
T.~Elsken, J.~H. Metzen, and F.~Hutter.
\newblock Neural architecture search: A survey.
\newblock {\em arXiv}, 2018.

\bibitem{fahlman1990cascade}
S.~E. Fahlman and C.~Lebiere.
\newblock The cascade-correlation learning architecture.
\newblock In {\em NIPS}, 1990.

\bibitem{feurer2015efficient}
M.~Feurer, A.~Klein, K.~Eggensperger, J.~Springenberg, M.~Blum, and F.~Hutter.
\newblock Efficient and robust automated machine learning.
\newblock In {\em NIPS}, 2015.

\bibitem{floreano2008neuroevolution}
D.~Floreano, P.~D{\"u}rr, and C.~Mattiussi.
\newblock Neuroevolution: from architectures to learning.
\newblock {\em Evolutionary Intelligence}, 2008.

\bibitem{goldberg1991comparative}
D.~E. Goldberg and K.~Deb.
\newblock A comparative analysis of selection schemes used in genetic
  algorithms.
\newblock {\em FOGA}, 1991.

\bibitem{he2016deep}
K.~He, X.~Zhang, S.~Ren, and J.~Sun.
\newblock Deep residual learning for image recognition.
\newblock In {\em CVPR}, 2016.

\bibitem{henderson2017deep}
P.~Henderson, R.~Islam, P.~Bachman, J.~Pineau, D.~Precup, and D.~Meger.
\newblock Deep reinforcement learning that matters.
\newblock {\em AAAI}, 2018.

\bibitem{hornby2006alps}
G.~S. Hornby.
\newblock Alps: the age-layered population structure for reducing the problem
  of premature convergence.
\newblock In {\em GECCO}, 2006.

\bibitem{hu2017squeeze}
J.~Hu, L.~Shen, and G.~Sun.
\newblock Squeeze-and-excitation networks.
\newblock {\em CVPR}, 2018.

\bibitem{huang2016densely}
G.~Huang, Z.~Liu, K.~Q. Weinberger, and L.~van~der Maaten.
\newblock Densely connected convolutional networks.
\newblock In {\em CVPR}, 2017.

\bibitem{huang2018gpipe}
Y.~Huang, Y.~Cheng, D.~Chen, H.~Lee, J.~Ngiam, Q.~V. Le, and Z.~Chen.
\newblock Gpipe: Efficient training of giant neural networks using pipeline
  parallelism.
\newblock {\em arXiv preprint arXiv:1811.06965}, 2018.

\bibitem{klein2017learning}
A.~Klein, S.~Falkner, J.~T. Springenberg, and F.~Hutter.
\newblock Learning curve prediction with bayesian neural networks.
\newblock {\em ICLR}, 2017.

\bibitem{krizhevsky2009learning}
A.~Krizhevsky and G.~Hinton.
\newblock Learning multiple layers of features from tiny images.
\newblock {\em Master's thesis, Dept. of Computer Science, U. of Toronto},
  2009.

\bibitem{krizhevsky2012imagenet}
A.~Krizhevsky, I.~Sutskever, and G.~E. Hinton.
\newblock Imagenet classification with deep convolutional neural networks.
\newblock In {\em NIPS}, 2012.

\bibitem{liu2017progressive}
C.~Liu, B.~Zoph, J.~Shlens, W.~Hua, L.-J. Li, L.~Fei-Fei, A.~Yuille, J.~Huang,
  and K.~Murphy.
\newblock Progressive neural architecture search.
\newblock {\em ECCV}, 2018.

\bibitem{liu2017hierarchical}
H.~Liu, K.~Simonyan, O.~Vinyals, C.~Fernando, and K.~Kavukcuoglu.
\newblock Hierarchical representations for efficient architecture search.
\newblock In {\em ICLR}, 2018.

\bibitem{mendoza2016towards}
H.~Mendoza, A.~Klein, M.~Feurer, J.~T. Springenberg, and F.~Hutter.
\newblock Towards automatically-tuned neural networks.
\newblock In {\em Workshop on Automatic Machine Learning}, 2016.

\bibitem{miikkulainen2017evolving}
R.~Miikkulainen, J.~Liang, E.~Meyerson, A.~Rawal, D.~Fink, O.~Francon, B.~Raju,
  A.~Navruzyan, N.~Duffy, and B.~Hodjat.
\newblock Evolving deep neural networks.
\newblock {\em arXiv}, 2017.

\bibitem{miller1989designing}
G.~F. Miller, P.~M. Todd, and S.~U. Hegde.
\newblock Designing neural networks using genetic algorithms.
\newblock In {\em ICGA}, 1989.

\bibitem{negrinho2017deeparchitect}
R.~Negrinho and G.~Gordon.
\newblock Deeparchitect: Automatically designing and training deep
  architectures.
\newblock {\em arXiv}, 2017.

\bibitem{pham2018faster}
H.~Pham, M.~Y. Guan, B.~Zoph, Q.~V. Le, and J.~Dean.
\newblock Faster discovery of neural architectures by searching for paths in a
  large model.
\newblock {\em ICLR Workshop}, 2018.

\bibitem{real2017large}
E.~Real, S.~Moore, A.~Selle, S.~Saxena, Y.~L. Suematsu, Q.~Le, and A.~Kurakin.
\newblock Large-scale evolution of image classifiers.
\newblock In {\em ICML}, 2017.

\bibitem{salimans2017evolution}
T.~Salimans, J.~Ho, X.~Chen, and I.~Sutskever.
\newblock Evolution strategies as a scalable alternative to reinforcement
  learning.
\newblock {\em arXiv}, 2017.

\bibitem{saxena2016convolutional}
S.~Saxena and J.~Verbeek.
\newblock Convolutional neural fabrics.
\newblock In {\em NIPS}, 2016.

\bibitem{simmons2011false}
J.~P. Simmons, L.~D. Nelson, and U.~Simonsohn.
\newblock False-positive psychology: Undisclosed flexibility in data collection
  and analysis allows presenting anything as significant.
\newblock {\em Psychological Science}, 2011.

\bibitem{srivastava2014dropout}
N.~Srivastava, G.~Hinton, A.~Krizhevsky, I.~Sutskever, and R.~Salakhutdinov.
\newblock Dropout: A simple way to prevent neural networks from overfitting.
\newblock {\em JMLR}, 2014.

\bibitem{stanley2005real}
K.~O. Stanley, B.~D. Bryant, and R.~Miikkulainen.
\newblock Real-time neuroevolution in the nero video game.
\newblock {\em TEVC}, 2005.

\bibitem{stanley2002evolving}
K.~O. Stanley and R.~Miikkulainen.
\newblock Evolving neural networks through augmenting topologies.
\newblock {\em Evol.\ Comput.}, 2002.

\bibitem{suganuma2017genetic}
M.~Suganuma, S.~Shirakawa, and T.~Nagao.
\newblock A genetic programming approach to designing convolutional neural
  network architectures.
\newblock In {\em GECCO}, 2017.

\bibitem{szegedy2017inception}
C.~Szegedy, S.~Ioffe, V.~Vanhoucke, and A.~A. Alemi.
\newblock Inception-v4, inception-resnet and the impact of residual connections
  on learning.
\newblock In {\em AAAI}, 2017.

\bibitem{szegedy2015going}
C.~Szegedy, W.~Liu, Y.~Jia, P.~Sermanet, S.~Reed, D.~Anguelov, D.~Erhan,
  V.~Vanhoucke, and A.~Rabinovich.
\newblock Going deeper with convolutions.
\newblock In {\em CVPR}, 2015.

\bibitem{wan2013regularization}
L.~Wan, M.~Zeiler, S.~Zhang, Y.~Le~Cun, and R.~Fergus.
\newblock Regularization of neural networks using dropconnect.
\newblock In {\em ICML}, 2013.

\bibitem{xie2017genetic}
L.~Xie and A.~Yuille.
\newblock Genetic {CNN}.
\newblock In {\em ICCV}, 2017.

\bibitem{xie2017aggregated}
S.~Xie, R.~Girshick, P.~Doll{\'a}r, Z.~Tu, and K.~He.
\newblock Aggregated residual transformations for deep neural networks.
\newblock In {\em CVPR}, 2017.

\bibitem{yao1999evolving}
X.~Yao.
\newblock Evolving artificial neural networks.
\newblock {\em IEEE}, 1999.

\bibitem{zagoruyko2016wide}
S.~Zagoruyko and N.~Komodakis.
\newblock Wide residual networks.
\newblock In {\em BMVC}, 2016.

\bibitem{zhang2017polynet}
X.~Zhang, Z.~Li, C.~C. Loy, and D.~Lin.
\newblock Polynet: A pursuit of structural diversity in very deep networks.
\newblock In {\em CVPR}, 2017.

\bibitem{zhong2017practical}
Z.~Zhong, J.~Yan, and C.-L. Liu.
\newblock Practical network blocks design with q-learning.
\newblock In {\em AAAI}, 2018.

\bibitem{zoph2016neural}
B.~Zoph and Q.~V. Le.
\newblock Neural architecture search with reinforcement learning.
\newblock In {\em ICLR}, 2016.

\bibitem{zoph2017learning}
B.~Zoph, V.~Vasudevan, J.~Shlens, and Q.~V. Le.
\newblock Learning transferable architectures for scalable image recognition.
\newblock In {\em CVPR}, 2018.

\end{thebibliography}

\renewcommand{\thesection}{A-\arabic{section}}
\renewcommand{\thefigure}{A-\arabic{figure}}
\setcounter{section}{0}
\setcounter{figure}{0}
\setcounter{table}{0}
\twocolumn[
\centerline{\textbf{\Large Supplement~A: Evolution and Reinforcement Learning}}
\vspace{20pt}
]

\begin{figure*}
\sbox0{\begin{subfigure}[b]{0.47\textwidth}
    \includegraphics[width=\linewidth,trim={23pt 30pt 70pt 20pt},clip]{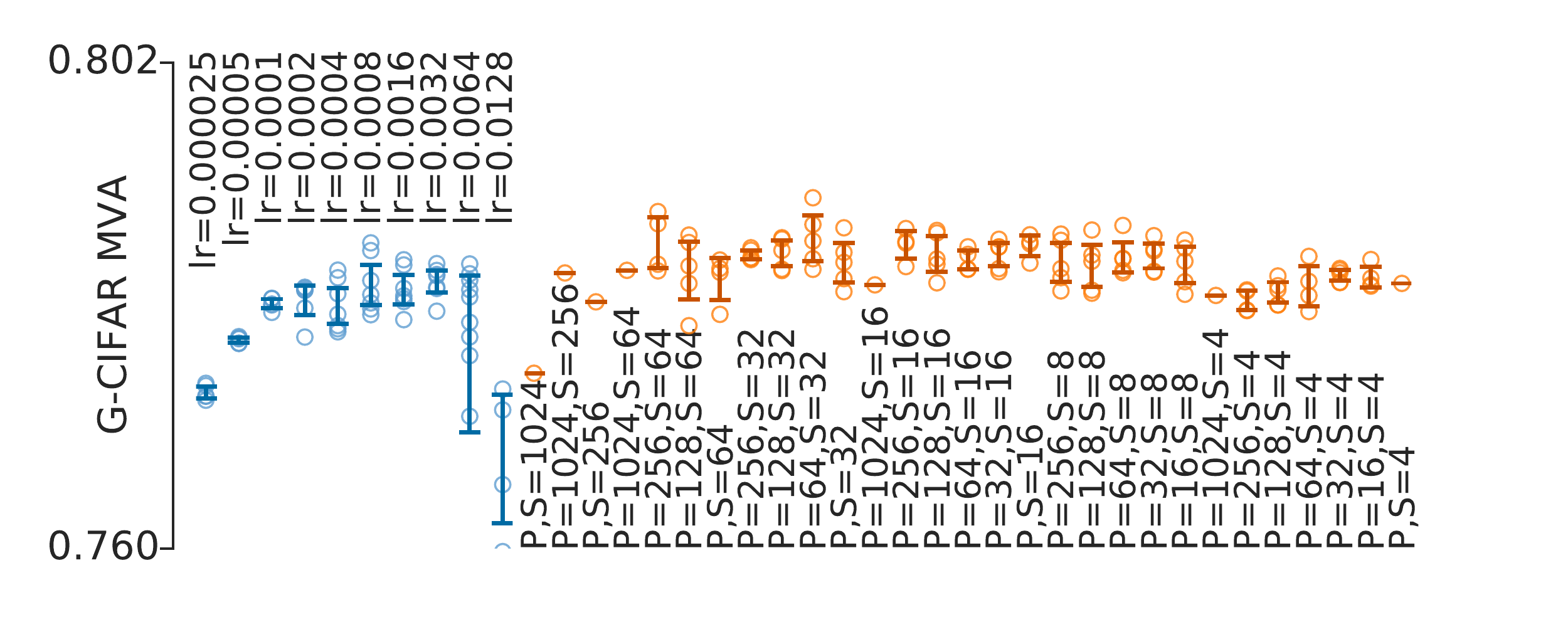}
    \caption{}
    \label{small_metaparams_subfig}
\end{subfigure}}
\sbox1{\begin{subfigure}[b]{0.47\textwidth}
    \includegraphics[width=\linewidth]{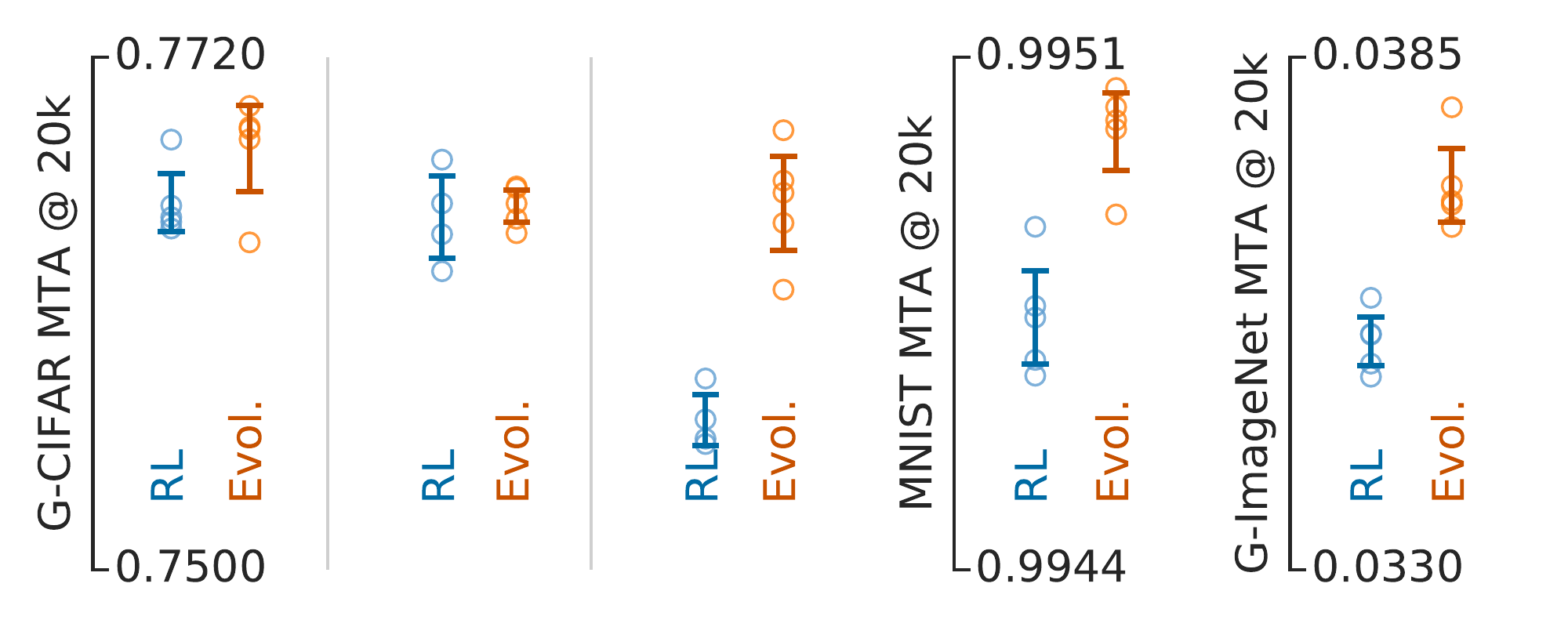}
    \caption{}
    \label{small_contexts20k_subfig}
\end{subfigure}}
\sbox2{\begin{subfigure}[b]{0.235\textwidth}
    \includegraphics[width=\linewidth]{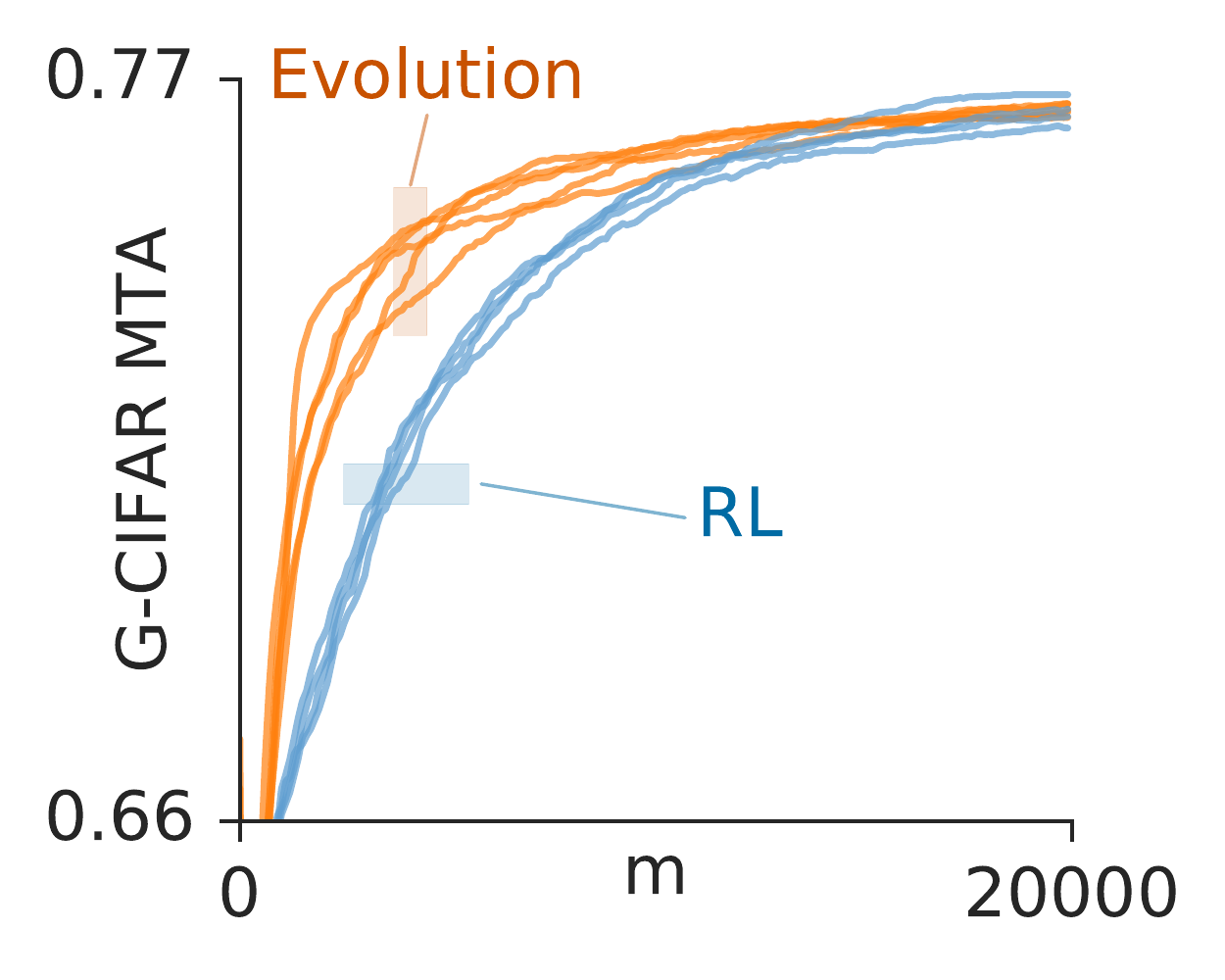}
    \caption{}
    \label{small_bigcell_subfig}
\end{subfigure}}
\sbox3{\begin{subfigure}[b]{0.235\textwidth}
    \includegraphics[width=\linewidth]{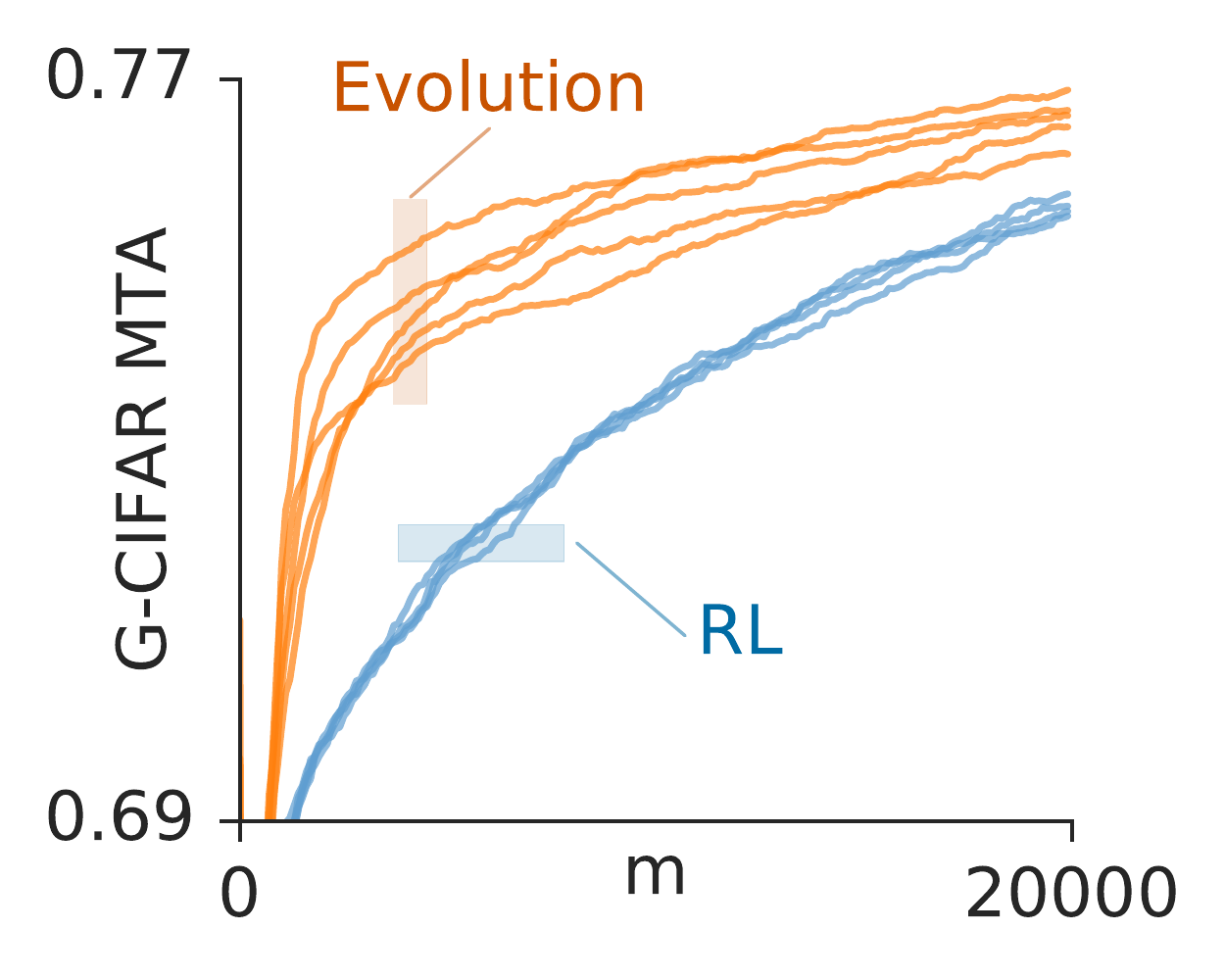}
    \caption{}
    \label{small_many_subfig}
\end{subfigure}}
\sbox4{\begin{subfigure}[b]{0.47\textwidth}
    \includegraphics[width=\linewidth]{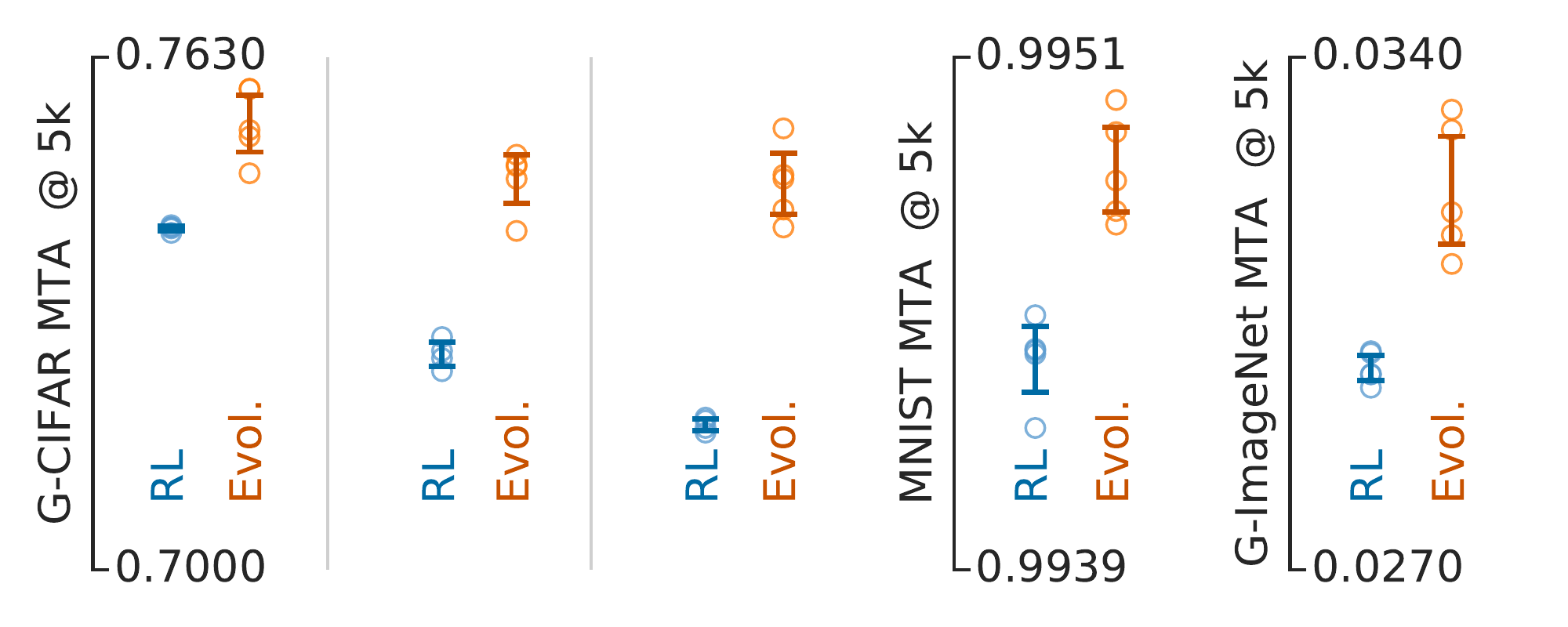}
    \caption{}
    \label{small_contexts5k_subfig}
\end{subfigure}}
\sbox5{\begin{subfigure}[b]{0.41\textwidth} 
    \includegraphics[width=\linewidth,height=1.4in]{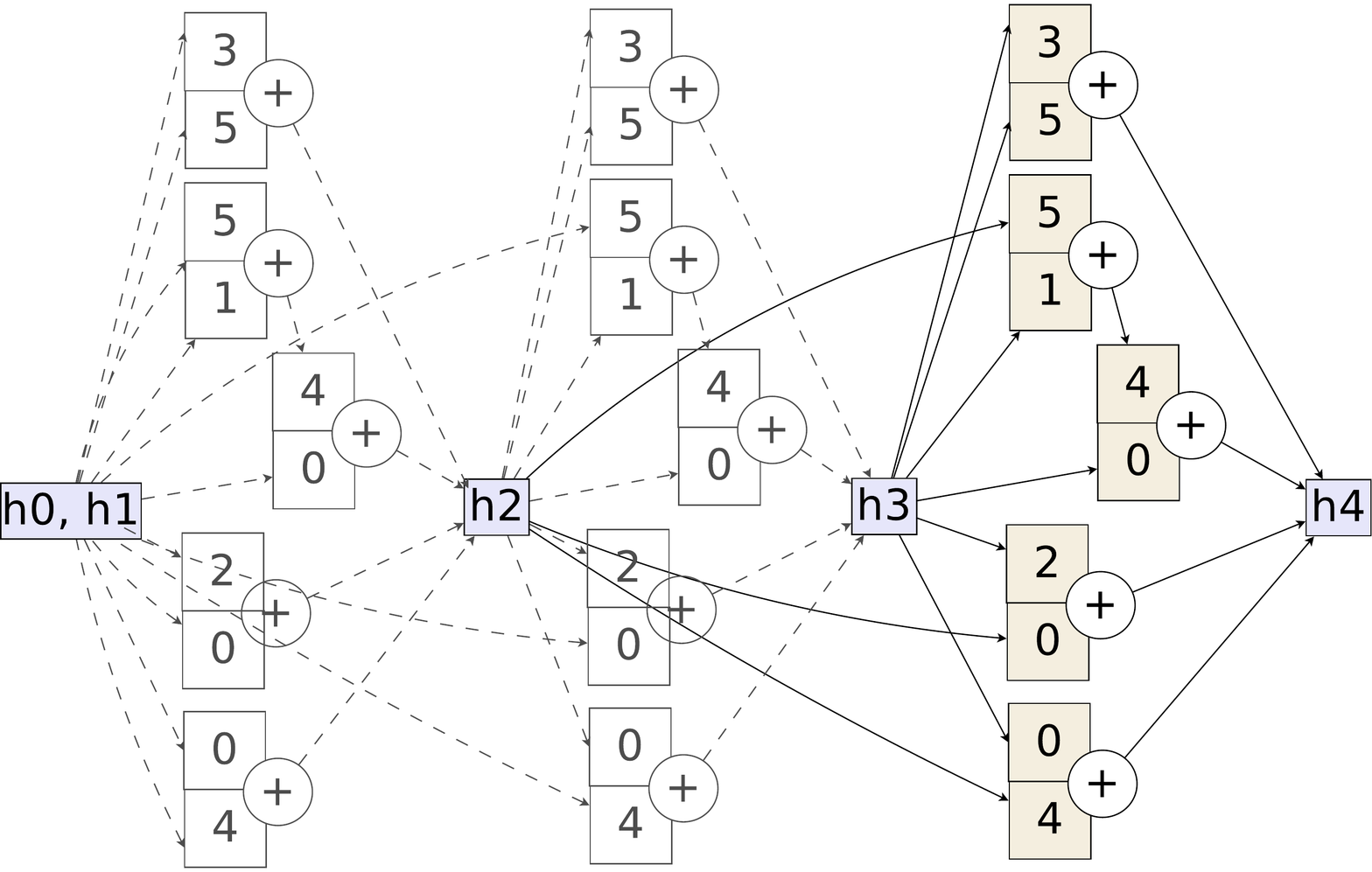}
    \caption{}
    \label{small_space_subfig}
\end{subfigure}}
\sbox6{
\begin{minipage}[b]{0.1\textwidth}
{\small
0 = sep. 3x3\\
1 = sep. 5x5\\
2 = sep. 7X7\\
3 = none\\
4 = avg. 3x3\\
5 = max 3x3\\
6 = dil. 3x3\\
7 = 1x7+7x1\\
\\
}
\end{minipage}}
\sbox7{\begin{subfigure}[b]{0.41\textwidth} 
    \includegraphics[width=\linewidth,height=1.4in]{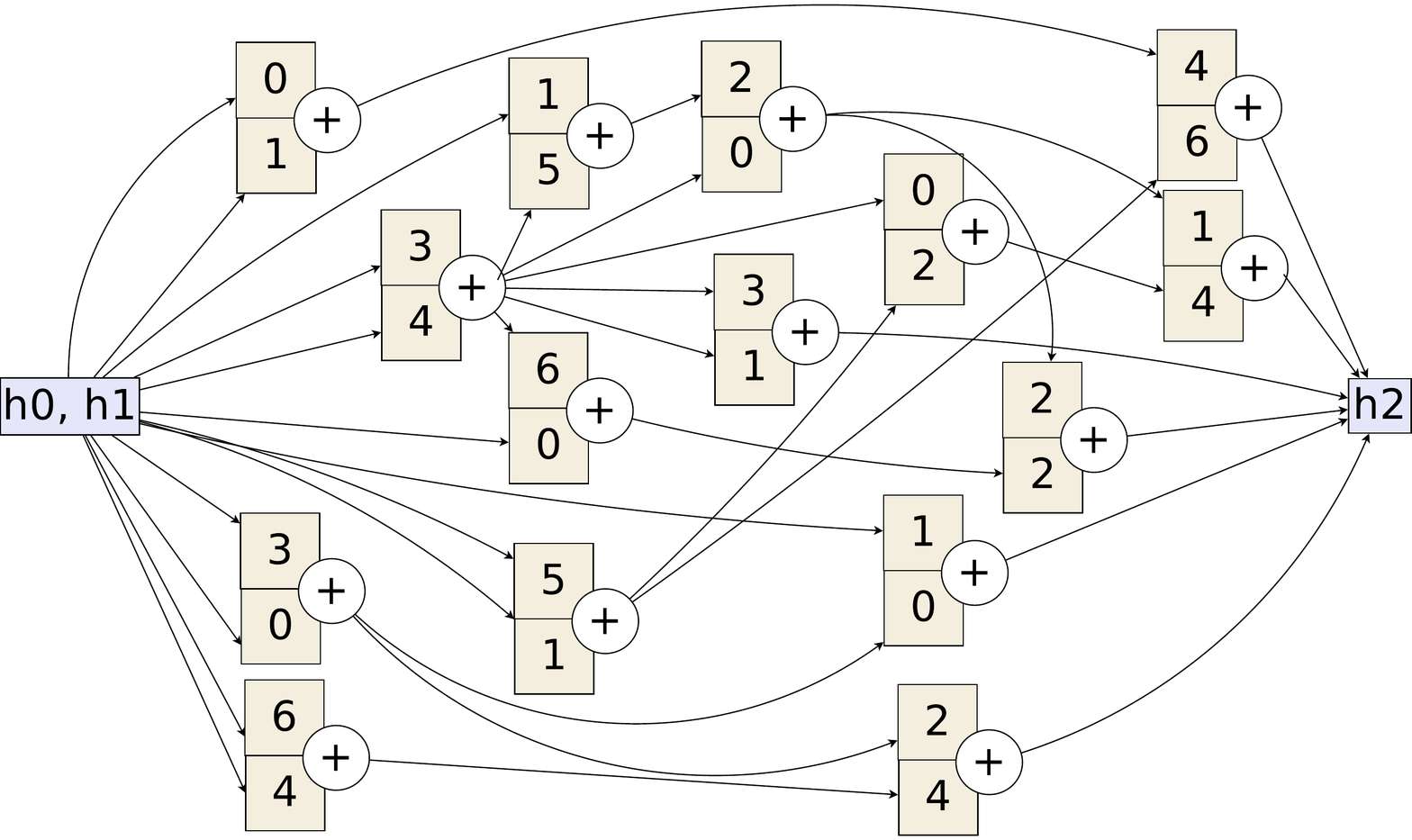}
    \caption{}
    \label{small_bigspace_subfig}
\end{subfigure}}
\begin{tabular}{cccc}
\multicolumn{2}{c}{\usebox0} & \multicolumn{2}{c}{\usebox1} \\
\usebox2 & \usebox3 & \multicolumn{2}{c}{\usebox4} \\
\end{tabular}
\begin{tabular}{ccc}
\usebox5 & \usebox6 & \usebox7
\end{tabular}
\caption{Evolution and RL in different contexts. Plots show repeated evolution (orange) and RL (blue) experiments side-by-side.
\textbf{(\subref{small_metaparams_subfig})} Summary of hyper-parameter optimization experiments on G-CIFAR. We swept the learning rate (lr) for RL (left) and the population size (P) and sample size (S) for evolution (right). We ran 5 experiments (circles) for each scenario. The vertical axis measures the mean validation accuracy (MVA) of the top 100 models in an experiment. Superposed on the raw data are $\pm \, 2 \, \textnormal{SEM}$ error bars. From these results, we selected best meta-parameters to use in the remainder of this figure.
\textbf{(\subref{small_contexts20k_subfig})} We assessed robustness by running the same experiments in 5 different contexts, spanning different datasets and search spaces: G-CIFAR/SP-I, G-CIFAR/SP-II, G-CIFAR/SP-III, MNIST/SP-I and G-ImageNet/SP-I, shown from left to right. These experiments ran to 20k models. The vertical axis measures the mean testing accuracy (MTA) of the top 100 models (selected by validation accuracy).
\textbf{(\subref{small_bigcell_subfig})} and \textbf{(\subref{small_many_subfig})} show a detailed view of the progress of the experiments in the G-CIFAR/SP-II and G-CIFAR/SP-III contexts, respectively. The horizontal axes indicate the number of models (m) produced as the experiment progresses.
\textbf{(\subref{small_contexts5k_subfig})} Resource-constrained settings may require stopping experiments early. At 5k models, evolution performs better than RL in all 5 contexts.
\textbf{(\subref{small_space_subfig})} and \textbf{(\subref{small_bigspace_subfig})} show a stack of normal cells of the best model found for G-CIFAR in the SP-I and SP-III search spaces, respectively. The ``h'' labels some of the hidden states. The ops (``avg 3x3'', \etc) are listed in full form in the text. Data flows from left to right. See the baseline study for a detailed description of these diagrams. In (\subref{small_space_subfig}), N=3 (see Methods section), so the cell is replicated three times; \ie the left two-thirds of the diagram (grayed out) are constrained to mirror the right third. This is in contrast with the vastly larger SP-III search space of (\subref{small_bigspace_subfig}), where a bigger, unconstrained construct without replication (N=1) is explored.}
\end{figure*}

\section{Motivation}

In this supplement, we will extend the comparison between evolution and reinforcement learning (RL) from the Results Section. Evolutionary algorithms and RL have been applied recently to the field of architecture search. Yet, comparison is difficult because studies tend to use novel search spaces, preventing direct attribution of the results to the algorithm. For example, the search space may be small instead of the algorithm being fast. The picture is blurred further by the use of different training techniques that affect model accuracy \cite{ciregan2012multi,wan2013regularization,srivastava2014dropout}, different definitions of \textit{FLOPs} that affect model compute cost\footnote{For example, see \url{https://stackoverflow.com/questions/329174/ what-is-flop-s-and-is-it-a-good-measure-of-performance}.} and different hardware platforms that affect algorithm run-time\footnote{A Tesla P100 can be twice as fast as a K40, for example.}. Accounting for all these factors, we will compare the two approaches in a variety of image classification contexts. To achieve statistical confidence, we will present repeated experiments without sampling bias. \nocite{simmons2011false}

\section{Setup}

All evolution and RL experiments used the NASNet search space design \cite{zoph2017learning}. Within this design, we define three concrete search spaces that differ in the number of pairwise combinations (C) and in the number of ops allowed (see Methods Section). In order of increasing size, we will refer to them as SP-I (\eg Figure~\ref{small_space_subfig}), SP-II, and SP-III (\eg Figure~\ref{small_bigspace_subfig}). SP-I is the exact variant used in the main text and in the study that we use as our baseline \cite{zoph2017learning}. SP-II increases the allowed ops from 8 to 19 (identity; 1x1 and 3x3 convs.; 3x3, 5x5 and 7x7 sep. convs.; 2x2 and 3x3 avg. pools; 2x2 min pool.; 2x2 and 3x3 max pools; 3x3, 5x5 and 7x7 dil. sep. convs.; 1x3 then 3x1 conv.; 1x7 then 7x1 conv.; 3x3 dil. conv. with rates 2, 4 and 6). SP-III allows for larger tree structures within the cells ($C$=$15$, same 19 ops).

The evolutionary algorithm is the same as that in the main text. The RL algorithm is the one used in the baseline study. We chose this baseline because, when we began, it had obtained the most accurate results on CIFAR-10, a popular dataset for image classifier architecture search.

We ran evolution and RL experiments for comparison purposes at different compute scales, always ensuring both approaches used identical conditions. In particular, evolution and RL used \textit{the same} code for network construction, training and evaluation. The experiments in this supplement were performed at a smaller compute scale than in the main text, to reduce resource usage: we used gray-scale versions of popular datasets (\eg ``G-Imagenet'' instead of ImageNet), we ran on CPU instead of GPU and trained relatively small models (F=8, see Methods~Details in main text) for only 4 epochs. Where unstated, the experiments ran on SP-I and G-CIFAR.

\section{Findings}

We first optimized the meta-parameters for evolution and for RL by running experiments with each algorithm, repeatedly, under each condition (Figure~\ref{small_metaparams_subfig}). We then compared the algorithms in 5 different contexts by swapping the dataset or the search space (Figure~\ref{small_contexts20k_subfig}). Evolution was either better than or equal to RL, with statistical significance. The best contexts for evolution and for RL are shown in more detail in Figures~\ref{small_bigcell_subfig} and~\ref{small_many_subfig}, respectively. They show the progress of 5 repeats of each algorithm. The initial speed of evolution is noticeable, especially in the largest search space (SP-III). Figures~\ref{small_space_subfig} and~\ref{small_bigspace_subfig} illustrate the top architectures from SP-I and SP-III, respectively. Regardless of context, Figure~\ref{small_contexts5k_subfig} indicates that accuracy under evolution increases significantly faster than RL at the initial stage. This stage was not accelerated by higher RL learning rates.

\section{Outcome}

The main text provides a comparison between algorithms for image classifier architecture search in the context of the SP-I search space on CIFAR-10, at scale. This supplement extends those results, varying the dataset and the search space by running many small experiments, confirming the conclusions of the main text.

\FloatBarrier
\clearpage
\renewcommand{\thesection}{B-\arabic{section}}
\renewcommand{\thefigure}{B-\arabic{figure}}
\setcounter{section}{0}
\setcounter{figure}{0}
\setcounter{table}{0}
\twocolumn[
\vspace{10pt}
\centerline{\textbf{\Large Supplement~B: Aging and Non-Aging Evolution}}
\vspace{20pt}
]

\section{Motivation}

In this supplement, we will extend the comparison between aging evolution (AE) and standard tournament selection / non-aging evolution (NAE). As was described in the Methods Section, the evolutionary algorithm used in this paper keeps the population size constant by always removing the oldest model whenever a new one is added; we will refer to this algorithm as AE. A recent paper used a similar method but kept the population size constant by removing the worst model in each tournament \cite{real2017large}; we will refer to that algorithm as NAE. This supplement will show how these two algorithms compare in a variety of contexts.

\section{Setup}

The search spaces and datasets were the same as in Supplement~A.

\section{Findings}

\begin{figure}[ht]
\centering
\includegraphics[width=0.97\linewidth]{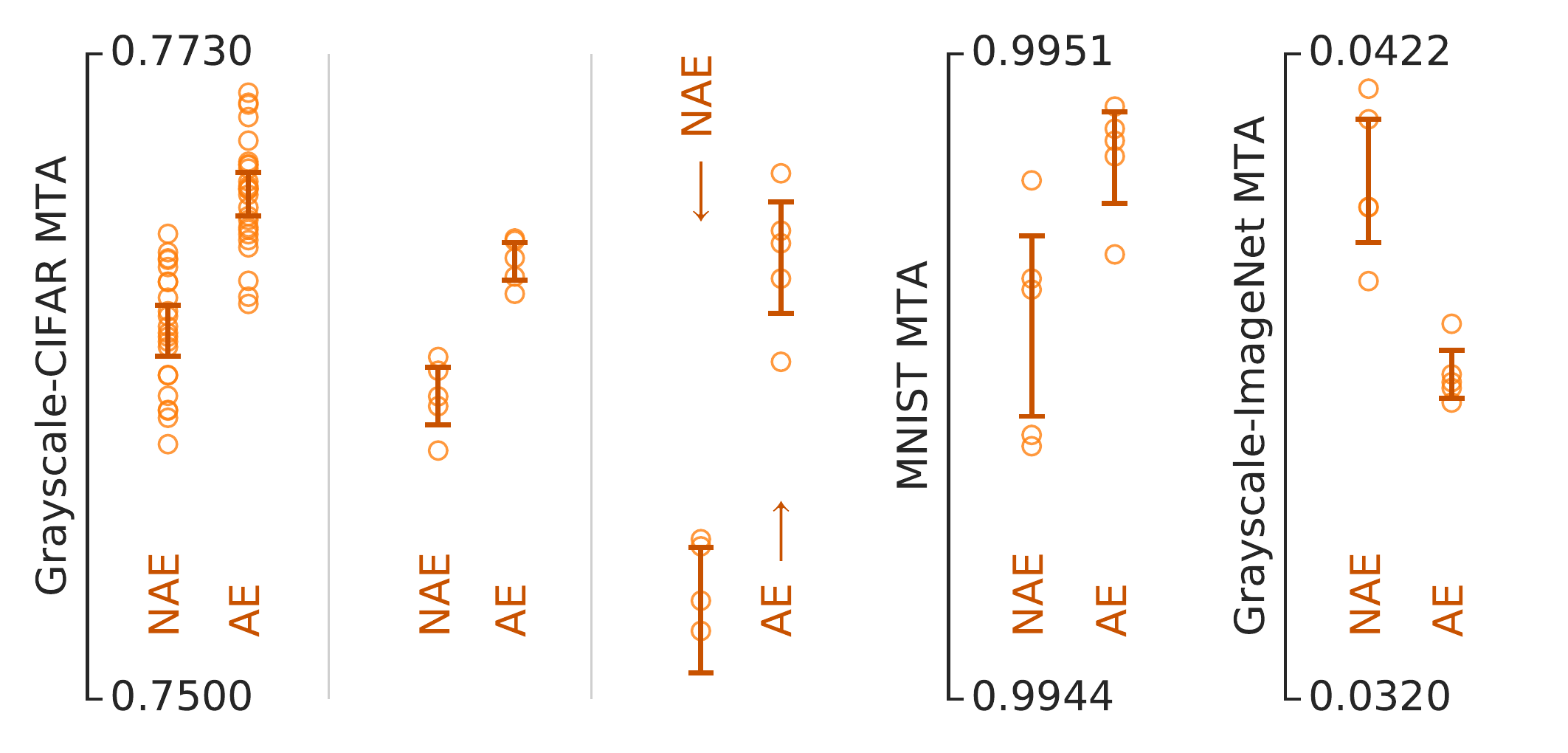}
\caption{
A comparison of NAE and AE under 5 different contexts, spanning different datasets and search spaces: G-CIFAR/SP-I, G-CIFAR/SP-II, G-CIFAR/SP-III, MNIST/SP-I and G-ImageNet/SP-I, shown from left to right. For each context, we show the final MTA of a few NAE and a few AE experiments (circles) in adjacent columns. We superpose $\pm \, 2 \, \textnormal{SEM}$ error bars, where SEM denotes the standard error of the mean. The first context contains many repeats with identical meta-parameters and their MTA values seem normally distributed (Shapiro--Wilks test). Under this normality assumption, the error bars represent 95\% confidence intervals.
}
\label{regularizing_contexts_subfig}
\end{figure}

We performed experiments in 5 different search space--dataset contexts. In each context, we ran several repeats of evolutionary search using NAE and AE (Figure \ref{regularizing_contexts_subfig}). Under 4 of the 5 contexts, AE resulted in statistically significant higher accuracy at the end of the runs, on average. The exception was the G-ImageNet search space, where the experiments were extremely short due to the compute demands of training on so much data using only CPUs. Interestingly, in the two contexts where the search space was bigger (SP-II and SP-III), \textit{all} AE runs did better than \textit{all} NAE runs.

\begin{figure}[ht]
\centering
\includegraphics[width=0.70\linewidth]{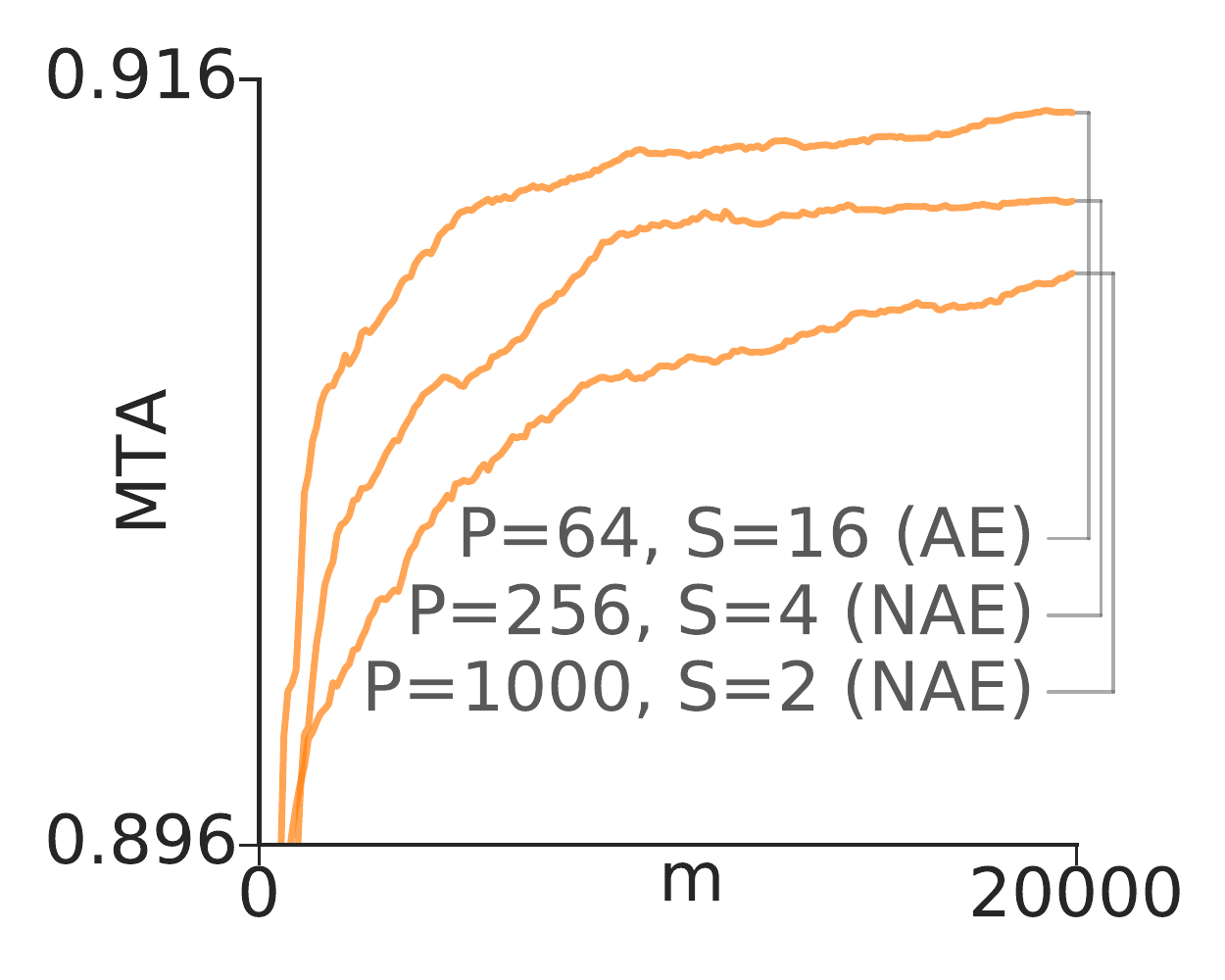}
\caption{
A comparison of AE and NAE at scale. These experiments use the same conditions as the main text (including dataset, search space, resources and duration). From top to bottom: an AE experiment with good AE meta-parameters from Supplement~A, an analogous NAE experiment, and an NAE experiment with the meta-parameters used in a recent study \cite{real2017large}. These accuracy values are not meaningful in absolute terms, as the models need to be augmented to reach their maximum accuracy, as described in the Methods Section).
}
\label{regularizing_large_fig}
\end{figure}

Additionally, we performed three experiments comparing AE and NAE at scale, under the same conditions as in the main text. The results, which can be seen in Figure~\ref{regularizing_large_fig}, provide some verification that observations from smaller CPU experiments in the previous paragraph generalize to the large-compute regime.

\section{Outcome}

The Discussion Section in the main text suggested that AE tends to perform better than NAE across various parameters for one fixed search space--dataset context. Such robustness is desirable for computationally demanding architecture search experiments, where we cannot always afford many runs to optimize the meta-parameters. This supplement extends those results to show that the conclusion holds across various contexts.

\FloatBarrier
\clearpage
\renewcommand{\thesection}{C-\arabic{section}}
\renewcommand{\thefigure}{C-\arabic{figure}}
\setcounter{section}{0}
\setcounter{figure}{0}
\setcounter{table}{0}
\twocolumn[
\vspace{10pt}
\centerline{\textbf{\Large Supplement~C: Aging Evolution in Toy Search Space}}
\vspace{20pt}
]

\section{Motivation}

As indicated in the Discussion Section, we suspect that aging may help navigate the noisy evaluation in an evolution experiment. We leave verification of this suspicion to future work, but for motivation we provide here a sanity check for it. We construct a toy search space in which the only difficulty is a noisy evaluation. Within this toy search space, we will see that aging evolution outperforms non-aging evolution.

\section{Setup}

The toy search space we use here does not involve any neural networks. The goal is to evolve solutions to a very simple, single-optimum, D-dimensional, noisy optimization problem with a signal-to-noise ratio matching that of our neuro-evolution experiments. 

The search space used is the set of vertices of a D-dimensional unit cube. A specific vertex is ``analogous'' to a neural network architecture in a real experiment. A vertex can be represented as a sequence of its coordinates (0s and 1s)---a bit-string. In other words, this bit-string constitutes a \textit{simulated architecture}. In a real experiment, training and evaluating an architecture yields a noisy accuracy. Likewise, in this toy search space, we assign a noisy \textit{simulated accuracy} (SA) to each cube vertex. The SA is the fraction of coordinates that are zero, plus a small amount of Gaussian noise ($\mu=0$, $\sigma=0.01$, matching the observed noise for neural networks). Thus, the goal is to get close to the optimum, the origin. The sample complexity used was 10k. This space is helpful because an experiment completes in milliseconds.

This optimization problem can be seen as a simplification of the evolutionary search for the minimum of a multi-dimensional integer-valued paraboloid with bounded support, where the mutations treat the values along each coordinate categorically. If we restrict the domain along each direction to the set \{0, 1\}, we reduce the problem to the unit cube described above. The paraboloid's value at a cube corner is just the number of coordinates that are not zero. We mention this connection because searching for the minimum of a paraboloid seems like a more natural choice for a trivial problem (``trivial'' compared to architecture search). The simpler unit cube version, however, was chosen because it permits faster computation.

We stress that these simulations are not intended to truly mimic architecture search experiments over the space of neural networks. We used them only as a testing ground for techniques that evolve solutions in the presence of noisy evaluations.

\section{Findings}

\begin{figure}[ht]
\centering
\includegraphics[width=0.70\linewidth]{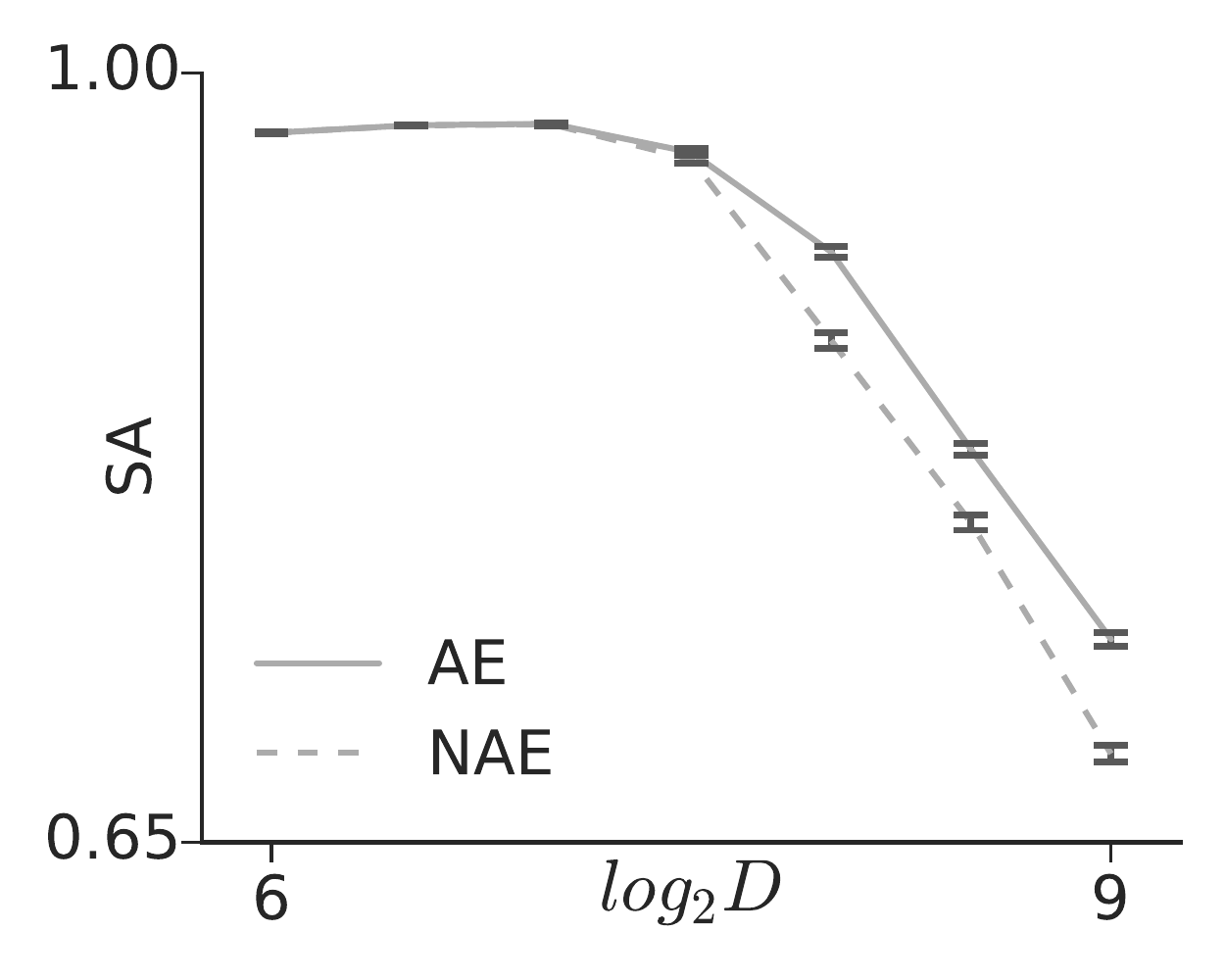}
\caption{
Results in the toy search space. The graph summarizes thousands of evolutionary search simulations. The vertical axis measures the simulated accuracy (SA) and the horizontal axis the dimensionality (D) of the problem, a measure of its difficulty. For each D, we optimized the meta-parameters for NAE and AE independently. To do this, we carried out 100 simulations for each meta-parameter combination and averaged the outcomes. We plot here the optima found, together with $\pm \, 2 \, \textnormal{SEM}$ error bars. The graph shows that in this toy search space, AE is never worse and is significantly better for larger D (note the broad range of the vertical axis).
}
\label{regularizing_sim_subfig}
\end{figure}

We found that optimized NAE and AE perform similarly in low-dimensional problems, which are easier. As the dimensionality (D) increases, AE becomes relatively better than NAE (Figure \ref{regularizing_sim_subfig}).

\section{Outcome}

The findings provide circumstantial evidence in favor of our suspicion that aging may help navigate noise (Discussion Section), suggesting that attempting to verify this with more generality may be an interesting direction for future work.

\FloatBarrier
\clearpage
\renewcommand{\thesection}{D-\arabic{section}}
\renewcommand{\thefigure}{D-\arabic{figure}}
\setcounter{section}{0}
\setcounter{figure}{0}
\setcounter{table}{0}
\twocolumn[
\vspace{10pt}
\centerline{\textbf{\Large Supplement~D: Additional AmoebaNets}}
\vspace{20pt}
]

\begin{figure*}[b]
\centering
\includegraphics[height=0.5\linewidth]{nasnet_space_outer_imagenet.pdf}
\hspace{0.03\linewidth}
\includegraphics[height=0.5\linewidth]{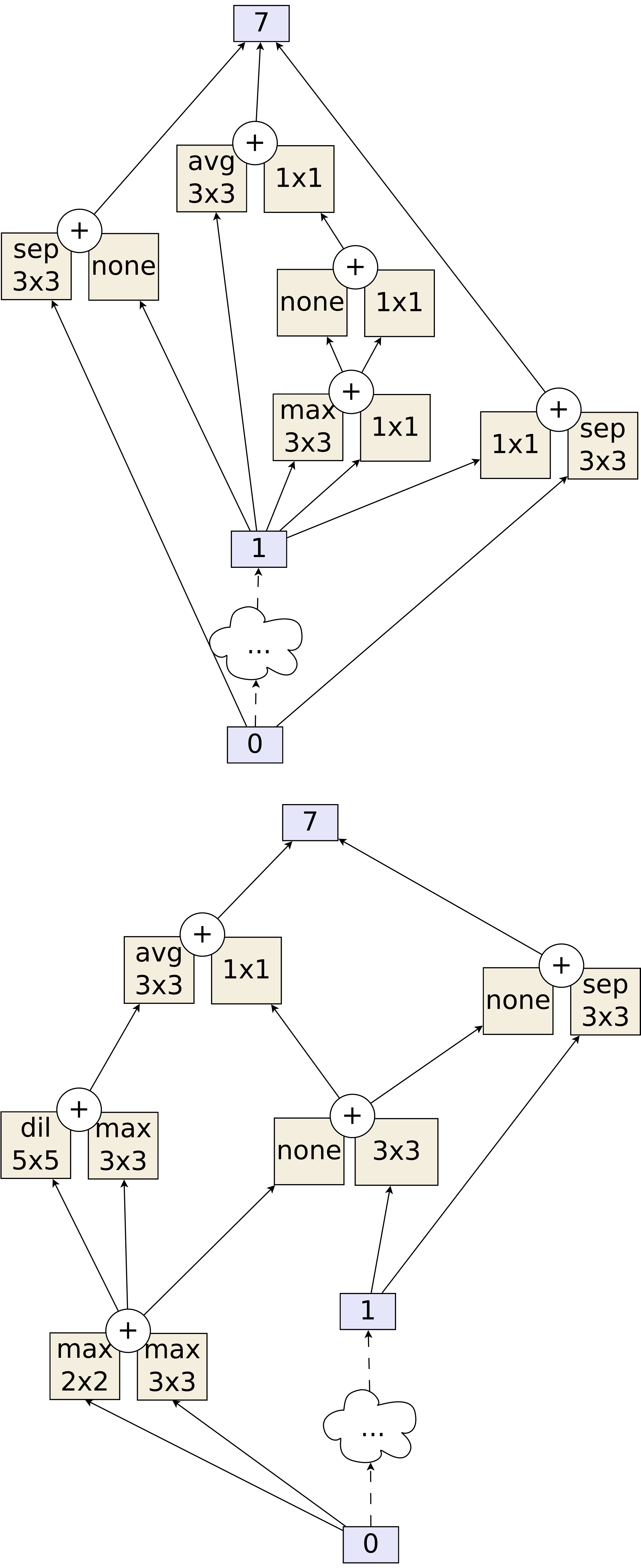}
\hspace{0.03\linewidth}
\includegraphics[height=0.5\linewidth]{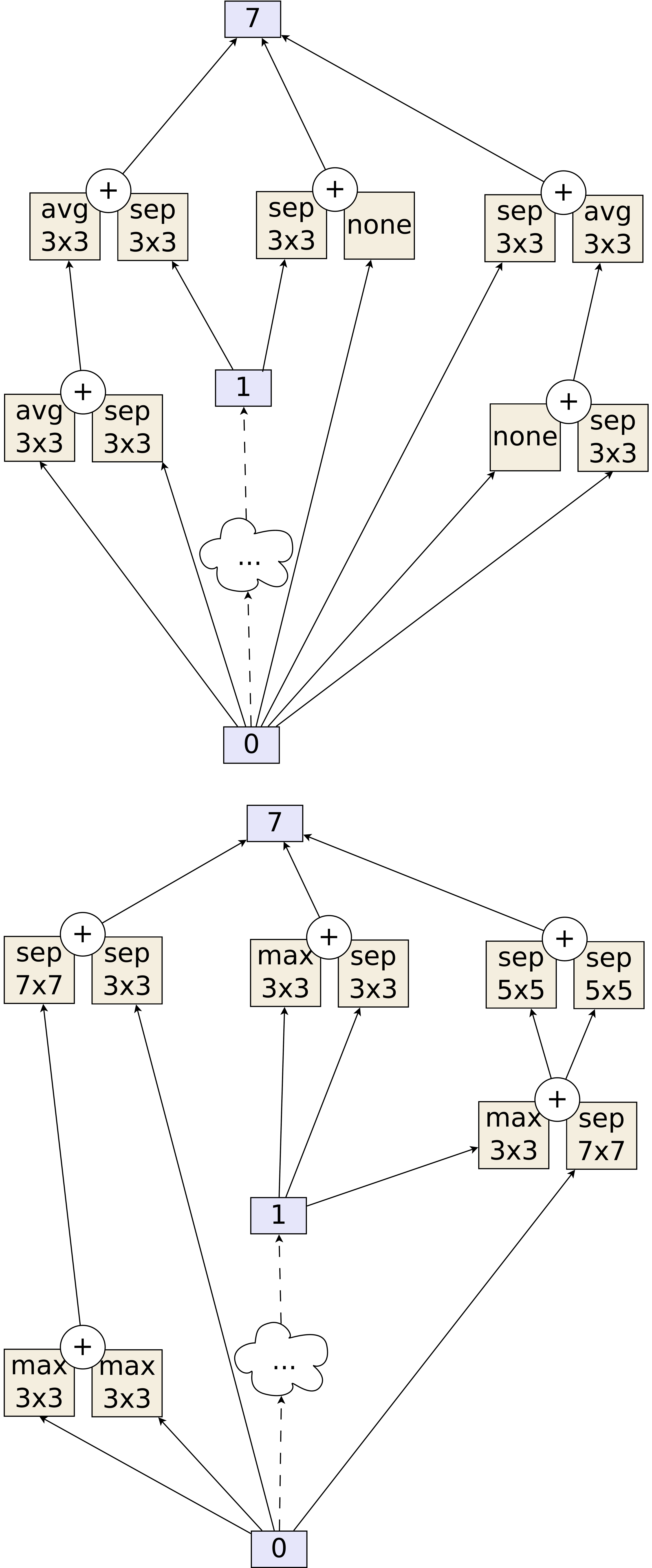}
\hspace{0.03\linewidth}
\includegraphics[height=0.5\linewidth]{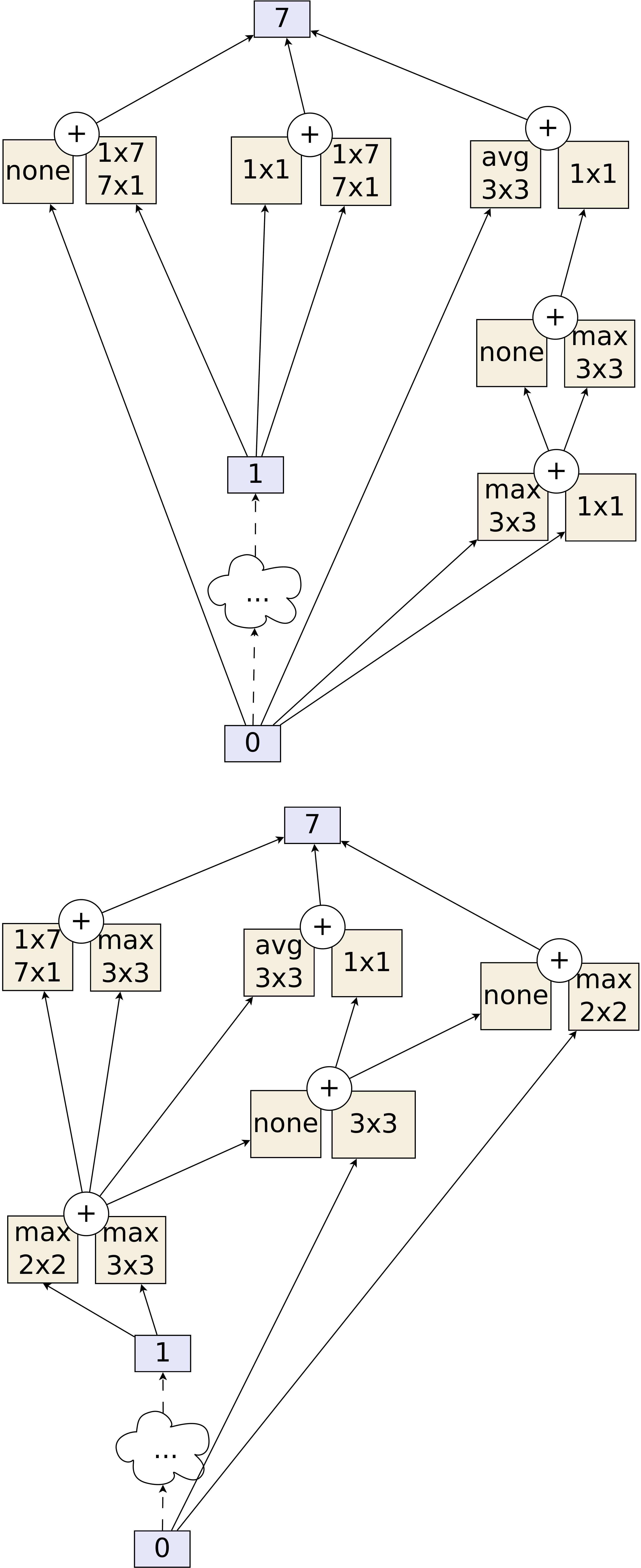}
\caption{Architectures of overall model and cells. From left to right: outline of the overall model \cite{zoph2017learning} and diagrams for the cell architectures discovered by evolution: \mbox{AmoebaNet-B}, \mbox{AmoebaNet-C}, and \mbox{AmoebaNet-D}. The three normal cells are on the top row and the three reduction cells are on the bottom row. The labeled activations or hidden states correspond to the cell inputs (``0'' and ``1'') and the cell output (``7'').}
\label{architectures_fig}
\end{figure*}

\section{Motivation}

In the Discussion Section, we briefly mentioned three additional models, \mbox{\textit{AmoebaNet-B}}, \mbox{\textit{AmoebaNet-C}}, and \mbox{\textit{AmoebaNet-D}}. While all three used the aging evolution algorithm presented the main text, there were some differences in the experimental setups: \mbox{AmoebaNet-B} was obtained through platform-aware architecture search; \mbox{AmoebaNet-C} was selected with a pareto-optimal criterion; and \mbox{AmoebaNet-D} involved multi-stage search, including manual extrapolation of the evolutionary process. Below we describe each of these models and the methods that produced them.

\section{Setup}

\mbox{\textbf{AmoebaNet-B}} was evolved by running experiments directly on Google TPUv2 hardware, since this was the target platform for its final evaluation. In the main text, the architecture had been discovered on GPU but the largest model was evaluated on TPUs. In contrast, here we perform the full process on TPUs. This \textit{architecture-aware} approach allows the evolutionary search to optimize even hardware-dependent aspects of the final accuracy, such as optimizations carried out by the compiler. The search setup was as in the main text, except that it used the larger SP-II space of Supplement~A and trained larger models (F=32) for longer (50 epochs). The selection of the top model was as follows. We picked from the experiment K=100 models. To do this, we binned the models by their number of parameters to cover the range, using B bins. From each bin, we took the top K/B models by validation accuracy. We then augmented all models to N=6 and F=32 and selected the one with the top validation accuracy.

\mbox{\textbf{AmoebaNet-C}} was discovered in the experiments described in the main text (see Methods and \mbox{Methods Details} sections). Instead of selecting the highest validation accuracy at the end of the experiments (as done in the main text), we picked a promising model while the experiments were still ongoing. This was done entirely for expediency, to be able to study a model while we waited for the search to complete. \mbox{AmoebaNet-C} was promising in that it stood out in a pareto-optimal sense: it was a high-accuracy outlier for its relatively small number of parameters. As opposed to all other architectures, AmoebaNet-C was selected based on CIFAR-10 test accuracy, because it was intended to only be benchmarked on ImageNet. This process was less methodical than the one used in the main text but because the model has been cited in the literature, we include it here for completeness.

\mbox{\textbf{AmoebaNet-D}} was obtained by manually modifying \mbox{AmoebaNet-B} by extrapolating evolution. To do this, we studied the progress of an experiment and identified which mutations were still causing improvements in fitness at the later stages of the process. By inspection, we found these mutations to be: replacing a 3x3 separable (sep.) convolution (conv.) with a 1x7 followed by 7x1 conv. in the normal cell, replacing a 5x5 sep. conv. by a 1x7 followed by 7x1 conv. in the reduction cell, and replacing a 3x3 sep. conv. with 3x3 avg. pool in the reduction cell. Additionally, we reduced the numeric precision from 32-bit to 16-bit floats, and set a learning rate schedule of step-wise decay, reducing by a factor of 0.88 every epoch. We trained for 35 epochs in total. To submit to Stanford DAWNBench (see Outcome section), we used N=2 and F=256.

\section{Findings}

Figure~\ref{architectures_fig} presents all three model architectures. We refrain from benchmarking these here. Instead, in the Outcome Section below, we will refer the reader to results presented elsewhere.

\section{Outcome}

In this supplement we have described additional evolutionary experiments that led to three new models. Such experiments were intended mainly to search for better models. Due to the resource-intensive nature of these methods, we forewent ablations and baselines in this supplement. For a more empirically rigorous approach, please refer to the process that produced \mbox{AmoebaNet-A} in the main text.

\mbox{\textbf{AmoebaNet-B}} had set a new state of the art on CIFAR-10 (2.13\% test error) in a previous preprint of this paper\footnote{Version 1 with same title on arXiv: \url{https://arxiv.org/pdf/1802.01548v1.pdf}} after being trained with cutout, but has since been superseded.

\mbox{\textbf{AmoebaNet-C}} had set the previous state-of-the-art top-1 accuracy on ImageNet after being trained with advanced data augmentation techniques in \cite{cubuk2018autoaugment}.

\mbox{\textbf{AmoebaNet-D}} won the Stanford DAWNBench competition for lowest training cost on ImageNet. The goal of this competition category was to minimize the monetary cost of training a model to 93\% top-5 accuracy. \mbox{AmoebaNet-D} costs \$49.30 to train. This was 16\% better than the second-best model, which was ResNet and which trained on the same hardware. The results were published in \cite{coleman2018analysis}.

\end{document}